\documentclass[nohyperref]{article}
\usepackage[table]{xcolor}

\usepackage{amsmath,amsfonts,bm, amsthm, mathrsfs, mathtools}

\newtheorem{lemma}{Lemma}

\numberwithin{theorem}{section}
\numberwithin{lemma}{section}
\numberwithin{proposition}{section}
\numberwithin{corollary}{section}

\def\ceil#1{\lceil #1 \rceil}

\def\1{\bm{1}}
\newcommand{\train}{\mathcal{D}}

\def\vw{{\bm{w}}}
\def\vx{{\bm{x}}}

\def\vz{{\bm{z}}}

\DeclareMathAlphabet{\mathsfit}{\encodingdefault}{\sfdefault}{m}{sl}
\SetMathAlphabet{\mathsfit}{bold}{\encodingdefault}{\sfdefault}{bx}{n}

\def\gD{{\mathcal{D}}}

\def\gL{{\mathcal{L}}}
\def\gM{{\mathcal{M}}}
\def\gN{{\mathcal{N}}}

\def\gW{{\mathcal{W}}}

\newcommand{\feat}{\vx}
\newcommand{\resp}{y}

\newcommand{\vocab}{V}
\newcommand{\real}{\mathbb{R}}
\newcommand{\temp}{\tau}
\newcommand{\ntest}{N_\ast}
\newcommand{\softmax}[3]{\frac{\text{exp}(\vw_{#1}^T\phi(#2; \gW) / #3)}{\sum_{v'=1}^{\vocab}\text{exp}(\vw_{v'}^T\phi(#2; \gW) / #3)}}
\newcommand{\elbo}[1]{\gL(#1)}
\newcommand{\telbo}[1]{\tilde{\gL}(#1)}
\newcommand{\param}{\bm\theta}
\newcommand{\pointmass}{\frac{1}{N_k}\sum_{n=1}^{N_k}\psi_{\param}(\phi(\feat_n^k))}
\newcommand{\batchpointmass}{\frac{1}{N_b}\sum_{n=1}^{N_b}\psi_{\param}(\phi(\feat_n^k))}
\newcommand{\modelname}{\textsc{thermometer}}
\newcommand{\llamaname}{LLaMA-2-Chat 7B}

\DeclareMathOperator*{\argmin}{arg\,min}

\usepackage{microtype,flushend}
\usepackage{graphicx}
\usepackage{subfigure}
\usepackage{booktabs} 

\usepackage[backref=page]{hyperref}
\usepackage{url}

\usepackage{amsfonts}       
\usepackage{mathrsfs}
\usepackage{nicefrac}       
\usepackage{enumitem}
\usepackage{quoting}
\usepackage{wrapfig}
\usepackage{makecell}
\usepackage{arydshln}
\usepackage{bbm}
\usepackage{dsfont}
\usepackage{caption} 
\usepackage{comment}
\usepackage[ruled,algo2e]{algorithm2e}
\usepackage{algorithm}
\usepackage{comment}
\usepackage{nicefrac}

\usepackage[accepted]{icml2024}
\hypersetup{
    colorlinks=true,
    citecolor =cyan,
    linkcolor=magenta,
    urlcolor=blue,
}

\usepackage{amsmath}
\usepackage{amssymb}
\usepackage{mathtools}
\usepackage{amsthm}
\usepackage{ifthen}
\usepackage{wrapfig}

\usepackage[capitalize,noabbrev]{cleveref}

\usepackage[textsize=tiny]{todonotes}

\icmltitlerunning{Thermometer: Towards Universal Calibration for LLMs}

\begin{document}
\newboolean{combinedfig}
\setboolean{combinedfig}{false} 

\twocolumn[
\icmltitle{Thermometer: Towards Universal Calibration for Large Language Models}


\icmlsetsymbol{equal}{*}

\begin{icmlauthorlist}
\icmlauthor{Maohao Shen}{mit}
\icmlauthor{Subhro Das}{ibm}
\icmlauthor{Kristjan Greenewald}{ibm}
\icmlauthor{Prasanna Sattigeri}{ibm}
\icmlauthor{Gregory Wornell}{mit}
\icmlauthor{Soumya Ghosh}{ibm}
\end{icmlauthorlist}

\icmlaffiliation{mit}{Department of Electrical Engineering and Computer Science, Massachusetts Institute of Technology, Cambridge, USA}
\icmlaffiliation{ibm}{MIT-IBM Watson AI Lab, IBM Research}
\icmlcorrespondingauthor{Maohao Shen}{maohao@mit.edu}
\icmlcorrespondingauthor{Soumya Ghosh}{ghoshso@us.ibm.com}

\icmlkeywords{Machine Learning, ICML}

\vskip 0.3in
]



\printAffiliationsAndNotice{}  

\begin{abstract}
We consider the issue of calibration in large language models (LLM). Recent studies have found that common interventions such as instruction tuning often result in poorly calibrated LLMs. Although calibration is well-explored in traditional applications, calibrating LLMs is uniquely challenging. These challenges stem as much from the severe computational requirements of LLMs as from their versatility, which allows them to be applied to diverse tasks. Addressing these challenges, we propose $\modelname$, a calibration approach tailored to LLMs. $\modelname$ learns an auxiliary model, given data from multiple tasks, for calibrating a LLM. It is computationally efficient, preserves the accuracy of the LLM, and produces better-calibrated responses for new tasks. Extensive empirical evaluations across various benchmarks demonstrate the effectiveness of the proposed method\footnote{The code is available at \url{https://github.com/maohaos2/Thermometer}.}.
\end{abstract}

\section{Introduction}  \label{sec:intro}
\begin{figure}[t]
    \centering
    \includegraphics[width=1\linewidth,clip]{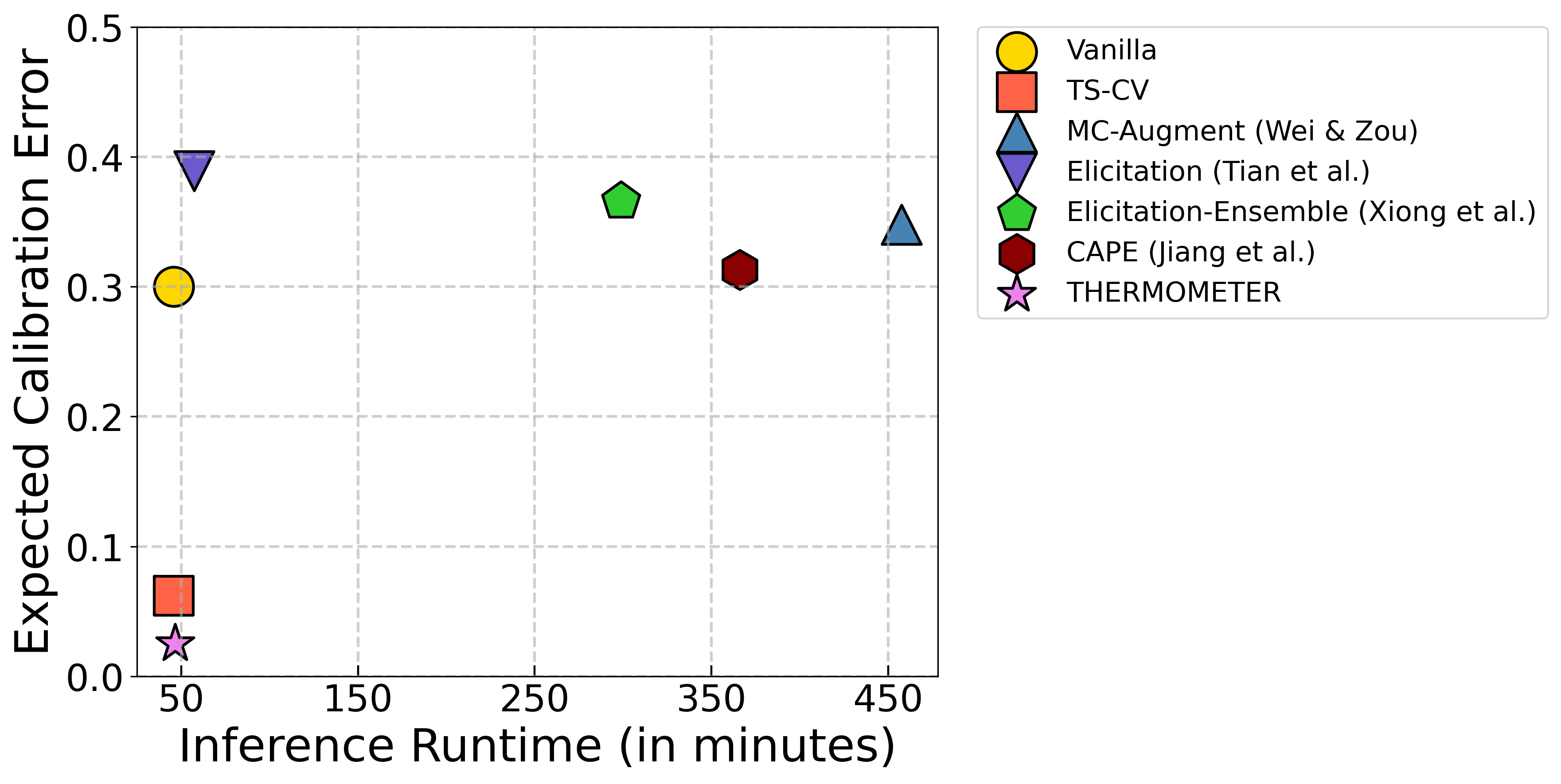}
\caption{\small{\textbf{Calibration performance against inference runtime.} Different methods for calibrating \llamaname~compared on the \textsc{Professional Law} task of MMLU on a A100, 40 GB GPU. The task contains $1533$ questions. Our method, $\modelname$ is significantly faster than methods that require multiple forward passes~\citep{wei-zou-2019-eda, xiong2023can, jiang2023calibrating} at inference time and achieves lower calibration error compared to methods with comparable runtime~\citep{tian2023just}. Vanilla refers to the no-calibration baseline, and TS-CV is a temperature scaling variant (\cref{sec:exp}). Similar trends hold for other benchmarks.}}
\label{figure:runtime}
\end{figure}

 Well-calibrated forecasts are a desirable property of any probabilistic forecaster. They ensure that probabilities produced by the forecaster can be interpreted as accurate confidence estimates of the forecasts. Informally, this implies that the forecaster is more often wrong on predictions made with low probabilities than those made with high probabilities. Such forecasts are useful for both enabling trust in the forecaster's predictions and incorporating the forecasts as part of a larger autonomous or semi-autonomous system. 

Large language models (LLMs)~\citep{brown2020language, raffel2020exploring, touvron2023llama} define probability distributions over sequences of tokens and produce probabilistic forecasts over future tokens in a sequence\footnote{and corresponding semantic entities represented by the tokens.}. Their ability to synthesize knowledge from large volumes of data, represented as sequences of tokens, has led to remarkable performance on diverse tasks, including question answering~\citep{hendrycks2020measuring}, commonsense reasoning~\citep{zhong2019improving}, and machine translation~\citep{zhu2020incorporating} among others. However, much like other probabilistic forecasters, before deploying LLMs in critical applications, it is important that they are well-calibrated in addition to being accurate. Unfortunately, a growing body of evidence suggests that while pre-trained LLMs are often well-calibrated, alignment interventions such as instruction tuning which make the pre-trained LLMs more usable, also harm calibration~\citep{openai2023gpt4, zhu2023calibration}. 

While calibration has long been studied~\citep{brier1950verification, dawid1982well, gneiting2007probabilistic} and different approaches for improving calibration properties of probabilistic forecasters exist~\citep{platt1999probabilistic, lakshminarayanan2017simple, gal2016dropout, guo2017calibration}, LLMs pose unique challenges. Training a LLM is expensive, and even inference typically incurs non-negligible expenses. This makes any calibration approach that requires multiple training runs prohibitively expensive. Even approaches that only require multiple inferences at test time can be unreasonably expensive for certain applications. Moreover, owing to their versatility, instruction-tuned LLMs are often applied, without further adaptation, to a diverse array of tasks. It is essential that methods for calibrating them do not 
affect the accuracy of the uncalibrated LLMs and that the calibration methods themselves can adapt to new tasks. Finally, measuring and alleviating calibration is challenging in cases where the LLM is required to generate free-form text. The equivalence class defined by the LLM-generated sequences that map to the same semantic content is large, making it challenging to assign meaningful confidence to the generation or even robustly assess the quality of the generation.  

We present $\modelname$ for calibrating LLMs while alleviating the above challenges. $\modelname$ learns, from multiple tasks, a parameterized mapping to map the outputs of the LLM to better-calibrated probabilities. Our approach is (i) computationally efficient: we do not require multiple training runs, and at inference time, we are only $\sim0.5\%$ slower than the uncalibrated LLM, (ii) accuracy preserving: we build on temperature scaling~\citep{guo2017calibration} which provably guarantees that greedy-decoded predictions do not change after our calibration procedure, and (iii) takes a step towards being universal: once trained we do not require retraining when exposed to a similar but new task. Moreover, similarly to recent work~\citep{kadavath2022language, lin2022teaching}, we circumvent the challenges posed by free-form text generation by mapping the free-form text generation task to a next-token prediction task. We empirically evaluate $\modelname$ on diverse benchmarks and models and find it consistently produce better-calibrated uncertainties than competing methods at a fraction of the computational cost. Moreover, we find $\modelname$ transfers across datasets and model scales. $\modelname$ trained for calibrating a smaller model (e.g., \llamaname) also improves calibration of larger models (e.g., LLaMA-2-Chat 70B) from the same family of models.

\section{Related Work}   \label{sec: related}
A variety of methods exist that aim to produce better-calibrated uncertainties. Post-hoc calibration methods learn to map the outputs of a pre-trained model to well-calibrated uncertainties. These include histogram binning~\citep{zadrozny2001obtaining, naeini2015}, isotonic regression~\citep{zadrozny2002transforming}, and approaches that assume parametric maps, including matrix, vector, and temperature scaling~\citep{platt1999probabilistic, guo2017calibration} as well as more ambitious variants~\citep{kull2019beyond} that aim to calibrate, in multi-class prediction problems, all class probabilities rather than just the probability of the most likely class (\emph{confidence calibration}). The next-token prediction task is also a multi-class prediction problem, albeit one with a large number of classes. Here, the number of classes equals the number of tokens in the LLM's vocabulary. This high dimensionality motivates us to focus on the more modest goal of confidence calibration. 

We learn an auxiliary model that, given an unlabeled dataset, predicts a dataset-specific temperature, allowing us to calibrate an LLM's uncertainties on a previously unseen task. Others~\citep{joy2023sample, yu2022robust} have also considered parameterized temperature maps but with a focus on more traditional vision models. \citet{joy2023sample} predict per-data instance temperatures, do not learn from multiple datasets, and find it necessary to learn a variational auto-encoder for representation learning jointly. In contrast, we leverage data from multiple tasks to generalize to new tasks, find no need for additional representation learning, and empirically find dataset-specific temperatures to better calibrate LLMs than data-instance-specific temperatures. \citet{yu2022robust} are motivated by making calibration robust to distribution shifts. Their approach involves a two-step procedure: independently learning dataset-specific temperatures for each dataset and then fitting a linear regression to the learned temperatures. In contrast, we jointly learn a non-linear auxiliary model across multiple datasets by simply maximizing a lower bound to the likelihood. Moreover, empirically, we find that $\modelname$ outperforms linear counterparts.

Other \emph{ab initio} approaches involve training with label-smoothing~\citep{szegedy2016rethinking}, mix-up augmentations~\citep{zhang2017mixup}, confidence penalties~\citep{pereyra2017regularizing}, focal loss~\citep{mukhoti2020calibrating}, or approximate Bayesian procedures~\citep{izmailov2021bayesian}. The substantial changes to the training process required by these approaches make them difficult to use with LLMs. Yet other approaches include ensembling over multiple models arrived at by retraining from different random initializations, for example, deep ensembles~\citep{lakshminarayanan2017simple}, or from perturbations of a single model, for example, Monte-Carlo dropout~\citep{gal2016dropout}. Training multiple LLM variants is prohibitively expensive, and while perturbations around a single model are possible, we find them to be not competitive with $\modelname$. 

\citet{jiang2021can, xiao2022uncertainty, chen2022close} empirically evaluate calibration of LLMs, find evidence of miscalibration, and evaluate existing calibration interventions and combinations thereof with varying degrees of success. \citet{park2022calibration} use mixup while \citet{desai2020} find temperature scaling and label smoothing effective for calibrating smaller encoder-only models. Others~\citep{lin2022teaching} employ supervised fine-tuning to produce verbalized uncertainties to be better calibrated on certain tasks. Such fine-tuning, however, is compute-intensive. An alternate body of work~\citep{zhang2021knowing, kadavath2022language, mielke2022reducing} approach calibration indirectly by learning an auxiliary model for predicting whether a generation is incorrect. There is also a body of work~\citep{abbas24a, han2023prototypical, jiang2023generative, zhao2021calibrate, zhou2024batch} that use the term calibration to mean de-biasing the predictions of a language model to biases introduced by the choice and ordering of in-context examples. This is distinct from our notion of statistical calibration. Yet others~\citep{tian2023just, xiong2023can}, similar to our approach, consider RLHF-tuned LLMs and find that they can express better-calibrated uncertainties with carefully crafted prompts. Our work is orthogonal, extending temperature scaling to tasks without labeled data, and can be combined with their approach.

\section{Background}   \label{sec: notation}
\textbf{Setup and Notation} We consider question answering tasks, both involving multiple-choice and free form answers. We pose this as a next token prediction problem by concatenating the question, any available context, and the multiple-choice options into a single prompt. In \cref{subsec:freeform}, we describe our treatment of free form answers within this setup. See \cref{app:prompts} for example prompts. Given the prompt, the next token predicted by the LLM is considered the answer to the question.  We denote the $n^{\text{th}}$ prompt in a dataset by $\feat_n$ and the corresponding completion by $\resp_n$. We assume that both the prompt and the completion have been tokenized, and the prompt $\feat_n = x_{n,t_n}, \ldots, x_{n,2}, x_{n,1}$ is a sequence of $t_n$ tokens. The tokens $x_{n,t}$ and $y_n$ take one of $\vocab$ values. We use $\train = \{\left(\feat_n, \resp_n\right)\}_{n=1}^{N}$ to denote a training set of prompt, completion pairs. We model 
$$p(\resp_n \mid \feat_n; \gM) = \frac{\text{exp}(\vw_{\resp_n}^T\phi(\vx_n; \gW))}{\sum_{v'=1}^{\vocab}\text{exp}(\vw_{v'}^T\phi(\vx_n; \gW))},$$
using a large language model, $\gM$, parameterized by weights, $\{\gW, \vw_1, \ldots, \vw_\vocab\}$. We use $p(\resp_n \mid \feat_n; \gM)$ as a notationally convenient stand in for ${p(Y = \resp_n \vert X = \feat_n; \gM)}$, where $Y$ and $X$ are random variables corresponding to the completion and the prompt. We also note that ${p(Y \mid X=\feat_n; \gM) \in \Delta^V}$ where $\Delta^V$ is a $V$-dimensional probability simplex. We view $\phi(\vx_n, \gW)$ as a feature extractor that transforms the input prompt $\feat_n$ into a $D$-dimensional representation.  
For notational brevity, we will suppress the dependence on $\gW$ and use $\phi(\vx_n, \gW)$ and $\phi(\vx_n)$ interchangeably.

\textbf{Confidence calibration} We aim to confidence calibrate $\gM$, such that among the completions whose top probability is $\beta$ the accuracy is also $\beta$,
$$\mathrm{Pr}(Y = \hat{Y} \vert \hat{P} = \beta) = \beta, \text{for all } \beta \in [0, 1],$$ where $\hat{Y} \!=\! \mathrm{argmax\; } p(Y \vert X; \!\gM\!)$ and $\hat{P}\! = \!\mathrm{max\; } p(Y \vert X; \gM)$.
 
 One common notion of miscalibration is the expected calibration error (ECE)~\citep{naeini2015}, $\mathbb{E}\left[ \big|\mathrm{Pr}(Y = \hat{Y} \mid \hat{P} = \beta) - \beta \big| \right]$. An empirical estimate of the ECE can be computed by partitioning $N$ samples into $M$ bins based on the confidences $\mathrm{max\; } p(Y \;\vert\; \feat_n; \gM)$ predicted by the model $\gM$. Let $B_m$ denote the set of samples allocated to the $M$-th bin. Then the empirical ECE estimate is the weighted average over the difference in accuracy,\\ $\small\text{acc}(B_m) = 1/|B_m| \sum_{n \in B_m} \mathbf{1}\left[\resp_n = \mathrm{argmax\; } p(Y \vert\feat_n; \gM)\right], $ and the average confidence within each bin,  \\$\small\text{conf}(B_m) \!=\! 1/|B_m|\sum_{n \in B_m} \!\!\mathrm{max\; } p(Y \vert \feat_n; \!\gM\!),$ $$\text{ECE} = \sum_{m=1}^{M} \frac{|B_m|}{N} \bigg|\text{acc}(B_m)-\text{conf}(B_m)\bigg|.$$  

\textbf{Temperature Scaling}
A large ECE score indicates the need for calibration as the model's prediction is poorly calibrated on the given dataset. Temperature scaling~\citep{guo2017calibration, platt1999probabilistic} is a widely used post-hoc calibration technique that introduces a single positive, scalar parameter $\temp \in \real_+$, the temperature, and defines,
$$p(\resp_n\mid \feat_n; \tau, \gM) = \softmax{\resp_n}{\feat_n}{\temp}.$$
The parameter $\temp$ is typically learned by minimizing the negative log likelihood (NLL) on a \emph{labeled} held-out dataset, $\hat{\temp} = \argmin_{\temp \in \real_+} -\sum_{n=1}^{N^*} \log p(\resp_n\mid \feat_n; \temp, \gM),$ freezing all other model parameters. Since temperature scaling only scales probabilities, it does not change the class with the maximum probability and preserves accuracy. 
Recent work~\citep{chen2022close, xiao2022uncertainty, tian2023just} has found temperature scaling effective at calibrating LLMs. Besides, it does not require additional tuning of the LLM or multiple forward passes at inference time. However, the requirement of a held-out labeled dataset can be severely limiting for LLMs. LLMs are often used on new tasks, where no labeled data is available! 

To alleviate this challenge, we propose to bypass dataset specific optimization and instead learn to predict the optimal temperature for a previously unseen and \emph{unlabeled} dataset.

\section{$\modelname$}    \label{sec:method}
Consider a multi-task setting. Given $K$ tasks with labeled datasets $\{\gD_k\}_{k=1}^{K} = \{\{\left(\feat^k_n, \resp^k_n\right)\}_{n=1}^{N_k}\}_{k=1}^{K}$, our goal is to learn to infer task-specific temperature parameters, in order to accurately infer the temperature for an unseen task given only $N_*$ prompts, $\gD_* = \{\feat_n^*\}_{n=1}^{N_*}$ but not the corresponding completions. 
To this end, we propose a probabilistic model with task-specific latent variables, $\temp_k$, for modeling the conditional distribution of completions given a prompt,
\begin{equation}
\begin{split}
p(&\train_1, \ldots \train_K \mid \nu_0; \gM) \\&= \prod_{k=1}^K \int p(\temp_k \mid \nu_0) \prod_{n=1}^{N_k} p(\resp_n^k \mid \feat_n^k, \tau_k; \gM) d\tau_k,
\end{split}
\label{eq:therm_model}
\end{equation}
where $p(\resp_n^k \mid \feat_n^k, \tau_k; \gM) = \softmax{\resp_n^k}{\feat_n^k}{\temp_k}$,  $p(\temp_k \mid \nu_0)$ is a prior on $\temp_k$ for regularizing it away from zero and large positive values, and $\nu_0$ is the prior's hyper-parameters.

Our key idea is to use a recognition network~\citep{dayan1995helmholtz}, to infer the per-task latent temperatures. We share the recognition network across all tasks, allowing us to amortize computation across tasks and sidestep the need for task specific optimization for inferring $\temp_k$. Crucially, we design our recognition network to condition only on $\phi(\feat_n^k)$ and not on the completions $\resp_n^k$, allowing us to infer the temperature for unlabeled data $\train^*$ at test time. We dub our recognition network, $\modelname$, which, much like a real $\modelname$, is designed to estimate temperature, albeit in the context of a dataset.
\subsection{A Variational Lower-bound} \label{sec:VI}
\begin{figure}[th]
    \centering
    \includegraphics[width=1\linewidth,clip]{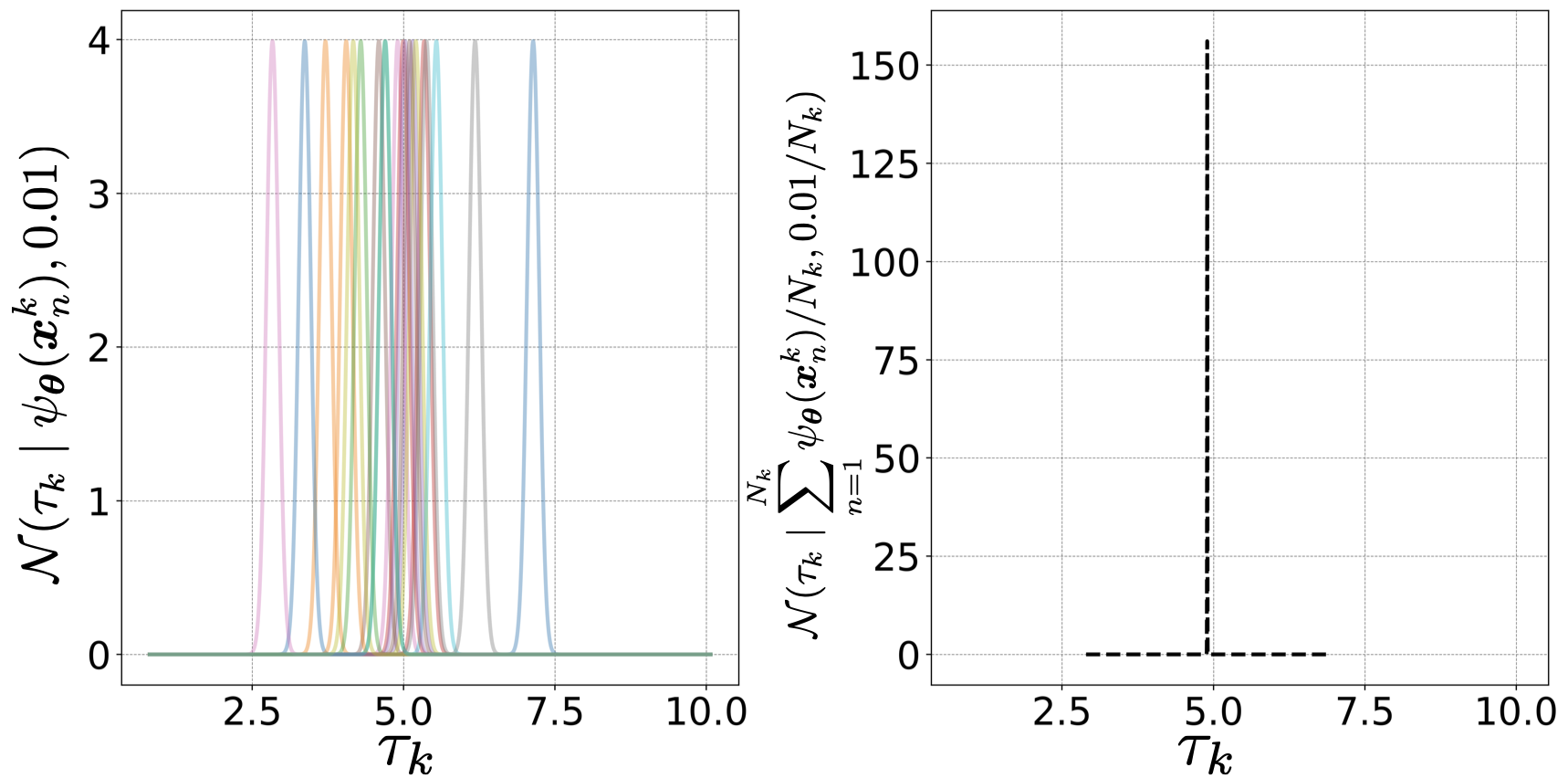}
\caption{\small{\textbf{Evidence accumulation.} We illustrate evidence accumulation for the \textsc{Professional Law} dataset from MMLU. It contains $N_k = 1533$ instances. The left panel, shows twenty of the possible $1533$ Gaussians, $\gN(\temp_k \mid \psi_{\param}(\phi(\feat_n^k)), \epsilon = 0.01)$. The right panel, plots the Gaussian proportional to $\prod_{n=1}^{1533}\gN(\temp_k \mid \psi_{\param}(\phi(\feat_n^k)), \epsilon = 0.01)$. The resulting Gaussian is nearly a point mass at $\frac{1}{1533}\sum_{n=1}^{1533}\psi_{\param}(\phi(\feat_n^k))$.}}
\label{figure:prod_gauss}
\end{figure}
 We work within the framework of variational inference, which provides us with a coherent objective to optimize for learning the recognition network. We begin by forming a variational lower bound of the logarithm of the marginal likelihood, $p(\train_1, \ldots \train_K \mid \nu_0; \gM)$, 
\begin{equation}
\begin{split}
\elbo{\param} =\sum_{k=1}^{K} &\mathbb{E}_{q(\temp_k;\param)}[\log p\left(\train_k \mid \temp_k; \gM \right)] \\&-\mathrm{KL}\left( q(\temp_k;\param) \;\vert\vert\; p(\temp_k\mid\nu_0) \right),
\end{split}
\label{eq: ELBO}
\end{equation}
where $q(\temp_k; \param)$ is a variational approximation to the posterior distribution $p(\temp_k \mid \train_1, \ldots, \train_k)$. We choose a product of Gaussian PDF for the variational approximation. Each corresponding Gaussian variable has a fixed variance $\epsilon$, and a mean parameterized by a recognition network (for example, a MLP), $\psi_{\param}: \real^D \rightarrow \real_{+}$, where $\param \in \real^P$ represents the parameters of the recognition network that are shared across $\train_1, \ldots, \train_k$. The resulting variational approximation is, 
\begin{equation}
\begin{split}
q(\temp_k; \param) &\propto \prod_{n=1}^{N_k} \gN(\temp_k \mid \psi_{\param}(\phi(\feat_n^k)), \epsilon),\\ &\gN\bigg(\temp_k \large\mid\frac{1}{N_k}\sum_{n=1}^{N_k}\psi_{\param}(\phi(\feat_n^k)), \frac{\epsilon}{N_{k}}\bigg),
\end{split}
\label{eq:vapprox}
\end{equation}
where the last equality follows from standard properties of Gaussian distributions. We provide a proof in \cref{app:proof2}. This particular choice of the variational parameterization affords us several benefits. First, it allows us to \emph{accumulate evidence}~\citep{bouchacourt2018multi} across data instances in $\train_k$, with the uncertainty in the inferred $\temp_k$ decreasing linearly with $N_k$. Second, by sharing $\param$ across datasets we share statistical strength across the training datasets $\train_1, \ldots, \train_K$ which allows us to use $\psi_{\param}$ to predict $\temp_*$ for an unseen dataset $\train_*$. Finally, when $\epsilon$ is small, our parameterization leads to a particularly simple algorithm.

\textbf{Training Objective}
Plugging in the variational approximation \cref{eq:vapprox} in \cref{eq: ELBO}, we have,
\begin{equation}
\elbo{\param} = \sum_{k=1}^{K} \mathbb{E}_{q(\temp_k;\param)}[\log p\left(\train_k \vert \temp_k; \gM\right) + \log p(\tau_k\vert \nu_0)] + \text{C}, 
\label{eq:celbo}
\end{equation}
where C\footnote{$\text{C} = K/2 + (K/2)\ln (2\pi\epsilon) $} is a constant independent of $\param$. By further observing that when $\epsilon$ is small, for example, $10^{-2}$, for even moderate $N_k$ the variational approximation $q(\temp_k \mid \param)$ is reasonably approximated by a point mass distribution $\delta(\tau_k - \pointmass)$ (\cref{figure:prod_gauss}). This leads to the objective for training $\modelname$,
\begin{equation}
\begin{split}
   \telbo{\param} = \sum_{k=1}^K &\log p\left(\train_k \;\bigg\vert\; \pointmass; \gM\right)\\ - &\lambda_{\text{reg}} \log p\left(\pointmass \;\bigg\vert\; \nu_0\right),
\end{split}
\end{equation}
where we have dropped the constant, plugged in the point mass approximation in \cref{eq:celbo}, and introduced a hyperparameter $\lambda_{\text{reg}}$ for balancing the prior and the likelihood terms. We place a Gamma prior, with a mode at one, on the temperatures $\temp_k$, $p(\temp_k \mid \nu_0) = \text{Gamma}(\temp_k \mid {\alpha_0=1.25, \beta_0=4})$\footnote{We use the shape scale parameterization, with shape = 1.25 and scale = 4.}. This prior enforces non-negative temperatures, and although the prior's mode at one encodes our belief that $\gM$ is already reasonably calibrated, the high prior variance emphasizes that this is only a weakly held belief. Finally, we train $\modelname$ by minimizing,
$\param^* = \underset{\param \in \real^P}{\argmin}  - \telbo{\param}$, or equivalently maximizing an approximate\footnote{the approximation is increasingly accurate with larger $N_k$.} lower bound to the log-marginal likelihood. We use stochastic optimization with mini-batched gradients to optimize $\telbo{\param}$. Algorithm \ref{alg:Alg1} summarizes our learning procedure. The temperature estimation step in the algorithm involves averaging over the minibatch which introduces a small bias in the minibatch gradients. However, this bias decreases linearly with increasing  batch size (\cref{lemm:Conc}), and disappears in the full-batch limit. Empirically, we find performance to be largely invariant to batch size (\cref{table:Ablation_train}).
\scalebox{0.9}
{
\begin{minipage}{1.1\linewidth}
\begin{algorithm}[H]
\SetAlgoLined

    \textbf{input}{ Training datasets of $K$ tasks $\{\gD_k\}_{k=1}^K$; pre-trained LLM $\gM$ with feature extractor $\phi$; prior hyper-parameters $\nu_0$; batch size $N_b$; learning rate $\gamma$; number of iterations $M$; checkpoint parameters $\{m', \mathrm{burnin}\}$; initialization $\param_{\text{init}}$ }
\\\\
$\triangleright$ Split $\{\gD_k\}_{k=1}^K$ into a development and validation set, $\mathrm{D}^{\mathrm{dev}} = \{\gD^{dev}_k\}_{k=1}^K, \mathrm{D}^{\mathrm{val}} = \{\gD^{val}_k\}_{k=1}^K$ 

$\triangleright$ Initialize $\modelname$, $\param_0 \leftarrow \param_{\text{init}}$; Denote the number of mini-batches by $B=\ceil{\frac{N}{N_b}}$, where $N = |\mathrm{D}^{\mathrm{dev}}|$.

\For{$m = 1,2, \ldots, M$}
{   
   $\triangleright$ Sample, uniformly at random, a task $k$ and its dataset $\gD^{dev}_k$.

   $\triangleright$ Sample a batch of $N_b$ samples from $\gD^{dev}_k$, i.e.,  $\{\left(\feat_n^k, \resp_n^k\right)\}_{n=1}^{N_b}$.  

   $\triangleright$ Estimate temperature $\hat{\temp}_k = \batchpointmass$.

   $\triangleright$ Evaluate the loss w.r.t batch of samples, i.e.,
    \small{\begin{align}
    -\telbo{\param}_{N_b} = -\sum_{n=1}^{N_b} \log p(\resp_n^k \mid \feat_n^k, \hat{\temp}_k ; \gM)
    - \frac{\lambda_{\text{reg}}}{B}\cdot\log p(\hat{\temp}_k \mid \nu_0) \nonumber
    \end{align}}

   $\triangleright$ Update the $\modelname$ parameter using AdamW, i.e., $\param_{m} \leftarrow \text{AdamW}(\param_{m-1}, \gamma, -\nabla_{\param}\telbo{\param}_b )$.

   \If {$m > \mathrm{burnin}$}{
   $\triangleright$ Checkpoint $\param_{m}$ every $m'$ iterations.
   }
}

$\triangleright$ Evaluate $\telbo{\param}$ on $\mathrm{D}^{\mathrm{val}}$ for each check-pointed $\param$ and set $\param^\ast$ to the check-pointed $\param$ with lowest $-\telbo{\param}$.  
\\\\
    \textbf{output}{ Optimized $\modelname$, $\param^\ast.$}

 \caption{$\modelname$ Training}\label{alg:variational}
 \label{alg:Alg1}
\end{algorithm}
\end{minipage}%
}
\subsection{Test Time Procedure} 
\label{subsec:testtime}
Given a new test task, we assume we have a collection of $\ntest$ inputs $\{\feat_n^{\ast}\}_{i=1}^{\ntest}$ with which we compute the temperature to be used for the task, 
\begin{equation}
\label{eq:test}
\temp_{*} = \frac{1}{\ntest} \sum_{n=1}^{\ntest}\psi_{\param^\ast}(\phi(\feat_n^\ast)).
\end{equation}
\textbf{Impact of test data size} 
At test time not all the data may be immediately available when computing the temperature $\temp_*$ (e.g. deploying on real-world tasks) and $\ntest$ in \eqref{eq:test}. As the empirical average used to compute $\temp_\ast$ is a proxy for $\mathbb{E}_{\feat\sim\mathcal{P}_\ast}[\psi_{\param}(\phi(\feat))]$, where $\mathcal{P}_\ast$ is the unknown data generating process for the test task, it is important that $\ntest$ not be too small or this approximation will not be accurate and performance may suffer.  %
We can control this error, however. Under an assumption that for fixed parameters, $\param$, $0 \leq \psi_{\param}(\phi(\feat)) \leq C_{\param}$ for some $C_{\param}$ (e.g. because the output of the $\modelname$ model must be bounded if the input is), we can apply a Bernstein inequality to obtain:
\begin{lemma}
    \label{lemm:Conc}
    Let $\mathcal{X}$ be the support set of the distribution of $\feat$, and assume we have $\ntest$ i.i.d. samples $\{\feat_n\}_{n=1}^{\ntest}$ from $\mathcal{P}_\ast$. Assume that for fixed parameters $\param$, $\sup_{\feat \in \mathcal{X}} \psi_{\param}(\phi(\feat)) \leq C_{\param}$ for some $C_{\param}$, and $\mathrm{Var}[\psi_{\param}(\phi(\feat_n))] \leq V_{\theta}$.\footnote{This can be replaced with $C^2_{\theta}/4$ if no additional knowledge of the variance is available.} Further assume that we are interested in measuring calibration error using a metric $\mathrm{CE}$ (for example, $\mathrm{ECE}$) that as a function of temperature, $\mathrm{CE}(\temp)$ is $L$-Lipschitz. Then with probability at least $1 - \frac{2}{{\ntest}^2}$,
    \begin{align*}
    &\mathbb{E} \left| \frac{1}{\ntest} \sum_{n=1}^{\ntest} \psi_{\param}(\phi(\feat_n)) - \mathbb{E}_{\feat\sim\mathcal{P}_\ast}\bigg[\psi_{\param}(\phi(\feat))\bigg]\right|\\
    &\qquad \leq  \frac{4}{3} C_{\theta} \frac{\log \ntest}{\ntest} + (2 V_{\theta})^{\frac{1}{2}} \sqrt{\frac{\log \ntest}{\ntest}}, \;\; \text{and,} \\
     &\mathrm{CE}\left(\frac{1}{\ntest} \sum_{n=1}^{n} \psi_{\param}(\phi(\feat_n))\right) - \mathrm{CE} \bigg(\mathbb{E}_{\feat\sim\mathcal{P}_\ast}\bigg[\psi_{\param}(\phi(\feat))\bigg]\bigg)
     \\
     &\qquad \leq L \left( \frac{4}{3} C_{\theta} \frac{\log \ntest}{\ntest} + (2 V_{\theta})^{\frac{1}{2}} \sqrt{\frac{\log \ntest}{\ntest}}\right).
    \end{align*}
\end{lemma}
The proof is in Appendix \ref{app:proof1}.
This result guarantees that these partial sums will converge quickly as $\ntest$ increases, and that any resulting loss of CE will be small. We confirm this guarantee experimentally in Appendix \ref{app:Ablation}. Our bound can be checked by practitioners as a guide to choosing $\ntest$, trading off accuracy and the need for more examples. In practice, both the Lipschitz constant and $V_{\theta}$ can be easily estimated. Specifically, the former is simple since the temperature is a one-dimensional last-layer parameter of an otherwise pretrained model, and the latter can be bounded by using an empirical estimate of the variance.\footnote{This can be done rigorously by deriving and adding a concentration upper bound for the error of the empirical variance before plugging into the bound in Lemma \ref{lemm:Conc} as $V_{\theta}$. }%
\subsection{Question Answering with Free Form Answers} \label{subsec:freeform}
As we introduced in Section~\ref{sec: notation}, given a sequence of tokens as prompts $\feat_n$, we aim to calibrate the LLM's prediction on its corresponding completion token $\resp_n$. In the context of a multiple-choice QA problem, $\resp_n$ typically represents a single token corresponding to one of the multiple-choice options, such as `A' or `B'.  However, in practice, LLMs are more commonly tasked with generating free-form responses for arbitrary questions without predefined answer options.

We address this issue by converting the free-form QA problem into a multiple-choice format, by posing a `Yes' or `No' question to an LLM, inquiring whether its generated response is correct or incorrect. Given a training set $\train = \{\left(\feat_n, \resp_n\right)\}_{n=1}^{N}$, where $\feat_n$ represents the question context, and $\resp_n$ denotes the true answers, we generate a new prompt $\tilde{\feat}_n$ by concatenating the original question $\feat_n$ with LLM's generated response $\vz_n$, i.e., $\tilde{\feat}_n = [\feat_n,\vz_n]$, and construct a new target $\tilde{\resp}_n$ by comparing generated response $\vz_n$ and true answer $\resp_n$, i.e., $\tilde{\resp}_n$ is assigned `Yes' if $\vz_n$ closely approximates $\resp_n$ based on a predefined similarity metric (for example, the Rouge-L score). The resulting calibration problem can be tackled in the same manner as multiple-choice QA, i.e., calibrate LLM's prediction of single completion token $\tilde{\resp}_n$ (`Yes' or `No') given the prompts $\tilde{\feat}_n$.

\begin{table*}[ht]
  \begin{center}
  \caption{\textbf{\llamaname~Average Calibration Performance on MMLU.} The results are reported as the mean and two standard error of the calibration results over 57 datasets. TS serves as the lower-bound as it has access to the labeled data of testing task.}
  \resizebox{2\columnwidth}{!}{%
  \begin{tabular}{cccccccc}
    \hline
    {\textbf{Methods}}  & \textbf{ECE} &  \textbf{TL-ECE}  & \textbf{MCE} & \textbf{NLL} & \textbf{Brier}\\
    \hline
    TS (lower-bound) & 0.062$\pm$0.008 & 0.123$\pm$0.011 & 0.298$\pm$0.051 & 1.205$\pm$0.043 & 0.161$\pm$0.006\\
    \hdashline
    Vanilla & 0.260$\pm$0.025 & 0.290$\pm$0.023 & 0.482$\pm$0.047 & 1.760$\pm$0.123 & 0.191$\pm$0.012\\
    TS-CV & 0.119$\pm$0.014 & 0.167$\pm$0.012 & 0.344$\pm$0.037 & 1.253$\pm$0.048 & 0.166$\pm$0.006  \\
    MC-Augment & 0.242$\pm$0.026 & 0.281$\pm$0.021 & 0.488$\pm$0.047 & 1.716$\pm$0.125 & 0.192$\pm$0.012 \\
    Elicitation & 0.315$\pm$0.016 & 0.340$\pm$0.014 &0.787 $\pm$0.034 & $\slash$ & $\slash$ \\
    Elicitation-Ensemble & 0.202$\pm$0.019 &0.255$\pm$0.017 &	0.544$\pm$0.055 & $\slash$ & $\slash$ \\
    CAPE & 0.196$\pm$0.021 & 0.247$\pm$0.016 &0.431$\pm$0.054 & 1.542$\pm$ 0.097 & 0.184$\pm$0.009 \\

    \rowcolor{lightgray}
    \textbf{\modelname} & \textbf{0.078$\pm$0.008} & \textbf{0.136$\pm$0.011} & \textbf{0.304$\pm$0.037} & \textbf{1.220$\pm$0.043} & \textbf{0.162$\pm$0.006} \\
   
    \hline
  \end{tabular}}
  \label{table:MMLU-LLaMA}
  \end{center}
\end{table*}

\begin{figure}[th]
    \centering
    \includegraphics[width=1\linewidth,clip]{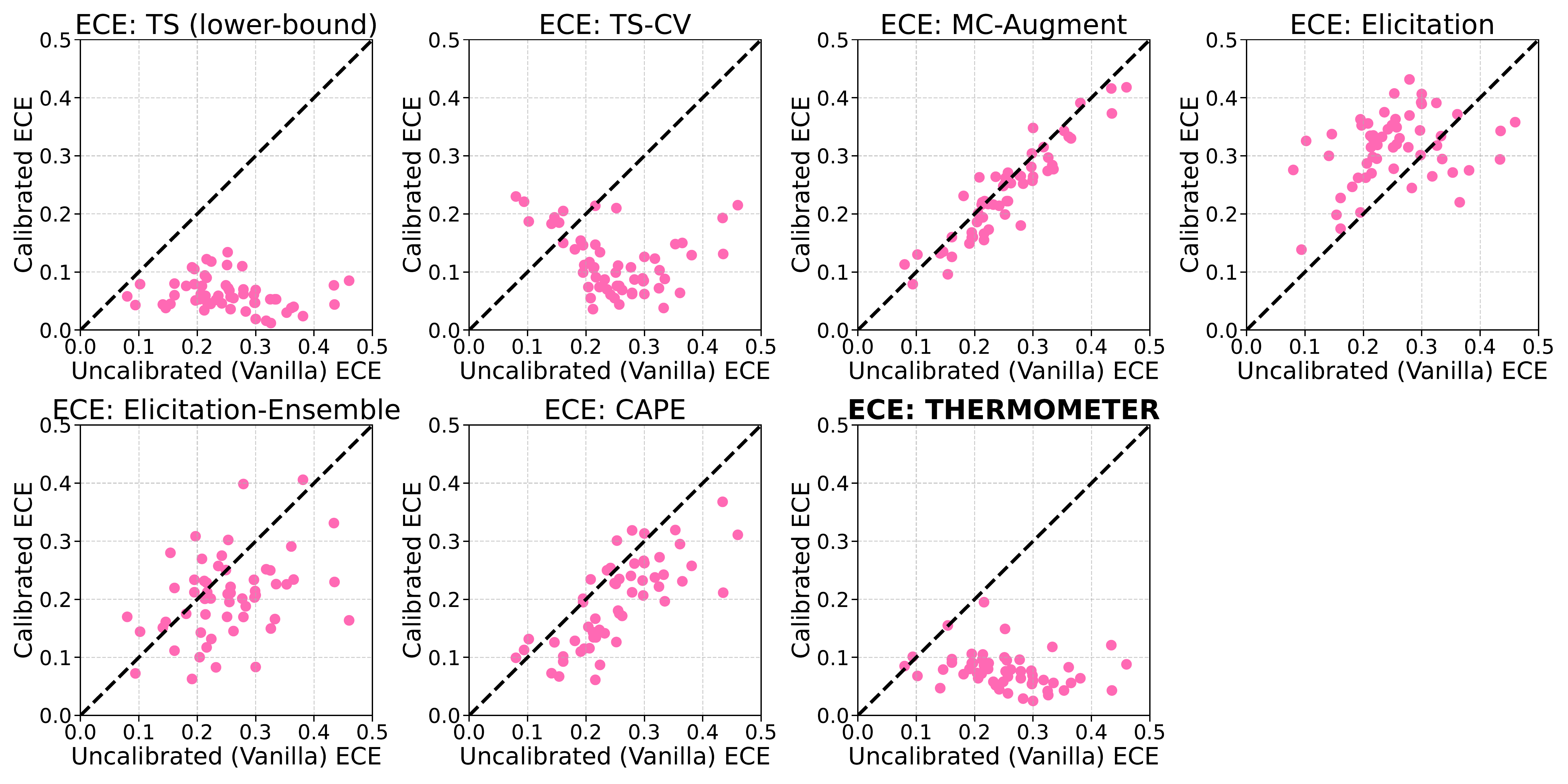}
    \vspace{-1em}
\caption{\small\textbf{\llamaname~Scatter Plots: ECE Score of 57 MMLU Datasets.} The x-axis and y-axis represent ECE score of uncalibared and calibrated model, respectively. $\modelname$ reduces calibration error on all 57 datasets, often substantially.}
\label{figure:MMLU-ECE-scatter-LLaMA}
\end{figure}

\section{Experiments}    \label{sec:exp}
We describe the experimental setup here, and present the main experimental results in \cref{exp:results}, and conduct a comprehensive analysis in \cref{exp:analysis}. Additional details and results are in \cref{app:results} and \cref{app:setup}.

\textbf{Benchmark Datasets}
We employ two widely used benchmark datasets for multiple-choice question-and-answer (QA) experiments: MMLU~\citep{hendrycks2020measuring} and BIG-bench~\citep{srivastava2022beyond}, and adopt MRQA~\citep{fisch2019mrqa} for experiments on QA with free form answers. MMLU contains datasets of exam questions from fifty seven subjects spanning various fields, each question comprises a question context and four possible answers. BIG-bench is a massive benchmark containing more than two hundred datasets covering a wide range of NLP tasks. We extracted multiple-choice QA datasets with over thousand training instances, which resulted in twenty three datasets. The MRQA benchmark comprises of a predefined train and development split, with six training datasets and six development datasets. The style of questions vary greatly both within and across the training and development sets. More details are in \cref{app:data}.
 
\textbf{Models}
We aim to enhance the calibration performance of instruction tuned language models. As decoder-only models become more ubiquitous in practice, our main experiments are conducted using decoder-only model \llamaname~with seven billion parameters~\citep{touvron2023llama}. To demonstrate the robust calibration performance of our proposed $\modelname$ with respect to different types of models, we also conduct a set of experiments using encoder-decoder model FLAN-T5-XL with three billion parameters~\citep{chung2022scaling}. $\modelname$ is an auxiliary model with a Multi-Layer Perceptron (MLP)-like structure described in \cref{app:model}.

\textbf{Evaluation Metrics}
We evaluate the calibration performance using following metrics: (1) Expected Calibration Error (ECE): measures the average error between prediction confidence and accuracy across different confidence intervals, for which we use 10 bins in our evaluation. (2) Maximum Calibration Error (MCE)~\citep{naeini2015}: identifies the largest error between prediction confidence and accuracy among all confidence bins, represents the worst-case calibration scenario. (3) Negative log likelihood (NLL): a proper scoring rule~\citep{gneiting2007strictly} that measures closeness to the data generating process in the Kullback-Leibler sense~\citep{deshpande2024are}. (4) Brier Score (Brier)~\citep{brier1950verification}: another proper scoring rule that measures the mean squared error between prediction confidence and one-hot labels. (5) Top Label ECE (TL-ECE)~\citep{gupta2021top}: measures the average error conditioned on both confidence and top-predicted labels. TL-ECE also serves as an upper bound of ECE, and is an alternate measure of calibration error.

\textbf{Implementation Details}
For multiple-choice QA tasks, we employ a leave-one-out approach to assess the $\modelname$'s calibration on unseen tasks. We train the model using $K-1$ datasets, and test the trained model on the remaining single testing task, repeating this process for all the $K$ datasets. For the QA tasks with free form answers, we use the established train and dev splits. We train $\modelname$ on MRQA's train split, and evaluate on the six held-out development datasets. In all experiments, we report results averaged over five random trials. See~\cref{app:implementation} for details.

\textbf{Baselines}
Our goal of calibrating models on tasks with no labeled data rules out many calibration approaches. Nonetheless, we compare $\modelname$ against several classical calibration methods as well as those recently proposed methods: (1) \textbf{Vanilla}: calibration performance of LLMs without any calibration techniques applied. (2) Temperature Scaling (\textbf{TS}): cheats by using labeled data from the testing task to tune the task-specific temperatures and establishes a lower bound for the  error. (3) Temperature Scaling with cross-validation (\textbf{TS-CV}): This variant of TS is similar to the setting of our approach. Here, the temperature is tuned using data from $K-1$ tasks, and used to calibrate the held-out task, without any task-specific adaptation. (4) Monte Carlo Dropout (\textbf{MC-Dropout})\footnote{We found dropout to not be functional in \llamaname, enabling dropout at inference time did not produced any variability, so we only report MC-Dropout comparisons with FLAN-T5-XL.}~\citep{gal2016dropout}: performs inference with random dropout multiple times, and averages the output probabilities across these inferences. (5) Monte Carlo Augmentation (\textbf{MC-Augment})~\citep{wei-zou-2019-eda}: employs random data augmentations of the prompts rather than dropout. (6) \textbf{Elicitation}~\citep{tian2023just} directly generates verbalized confidence by providing the LLMs with carefully crafted prompts. (7) \textbf{Elicitation-Ensemble}~\citep{xiong2023can} improves the quality of verbalized confidence by aggregation of multiple samples. (8) \textbf{CAPE}~\citep{jiang2023calibrating} performs prompt augmentation using option permutation.

\subsection{Main Results} \label{exp:results}
\begin{table*}[ht]
  \begin{center}
  \caption{\textbf{\llamaname~Average Calibration Performance on MRQA.} The results are reported as the mean and two standard error of the calibration results over the six MRQA validation datasets.}
   \resizebox{2\columnwidth}{!}{%
  \begin{tabular}{cccccccc}
    \hline
    \textbf{Methods}  & \textbf{ECE} &  \textbf{TL-ECE}  & \textbf{MCE} & \textbf{NLL} & \textbf{Brier}\\
    \hline
    TS (lower-bound) & 0.029$\pm$0.005 & 0.058$\pm$0.021 & 0.115$\pm$0.015 & 0.536$\pm$0.042 & 0.178$\pm$0.017\\
    \hdashline
    Vanilla & 0.127$\pm$0.026 & 0.146$\pm$0.023 & 0.198$\pm$0.023 & 0.656$\pm$0.068 & 0.198$\pm$0.022\\
    TS-CV & 0.071$\pm$0.011 & 0.099$\pm$0.014 & 0.157$\pm$0.015 & 0.556$\pm$0.042 & 0.183$\pm$0.017\\
    MC-Augment & 0.335$\pm$0.044 & 0.411$\pm$0.034 & 0.679$\pm$0.052 & 0.997$\pm$0.081 & 0.370$\pm$0.026\\
    Elicitation &0.130$\pm$0.031 &0.207$\pm$0.036	& 0.522$\pm$0.169 & $\slash$ & $\slash$ \\
    Elicitation-Ensemble & 0.171$\pm$0.047 &0.237$\pm$0.032	 & 0.552$\pm$0.206 & $\slash$ & $\slash$ \\
    CAPE & 0.067$\pm$0.016 &0.098$\pm$0.028 &\textbf{0.156$\pm$0.060} &  0.556$\pm$ 0.087& 0.183$\pm$0.036\\
    \rowcolor{lightgray}
    \textbf{\modelname} & \textbf{0.065$\pm$0.008} & \textbf{0.093$\pm$0.016} & \textbf{0.163$\pm$0.027} & \textbf{0.551$\pm$0.039} & \textbf{0.182$\pm$0.016} \\
    \hline
  \end{tabular}}
  \label{table:mrqa-LLaMA}
  \end{center}
\end{table*}

\begin{figure}[ht]
    \centering
   \includegraphics[width=0.6\linewidth,clip]{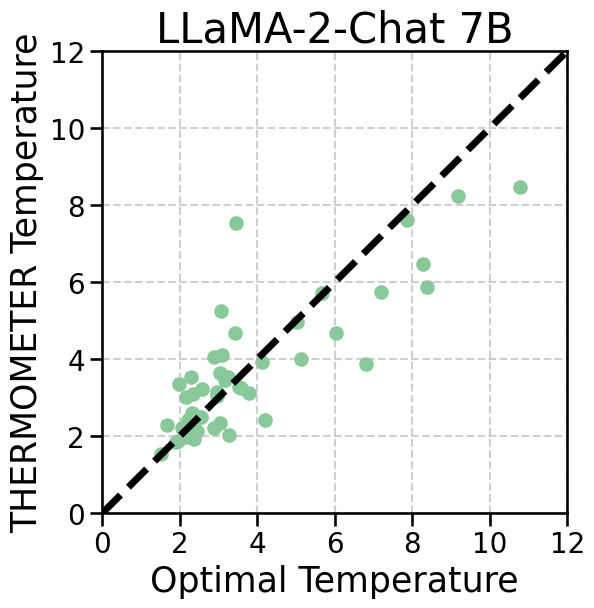}
   \vspace{-0.5em}
\caption{\textbf{$\modelname$ against temperature scaling on MMLU.} Comparison of $\modelname$ predictions and temperatures obtained by temperature scaling. Each green dot represents one MMLU task. The x-coordinate is the temperature learned via temperature scaling and the y-coordinate is the $\modelname$ predicted temperature. $\modelname$ accurately predicts the temperature for unseen task.}
\label{figure:MMLU-temperature}
\end{figure}

\begin{figure*}[ht]
    \centering
    \includegraphics[width=0.3\linewidth]{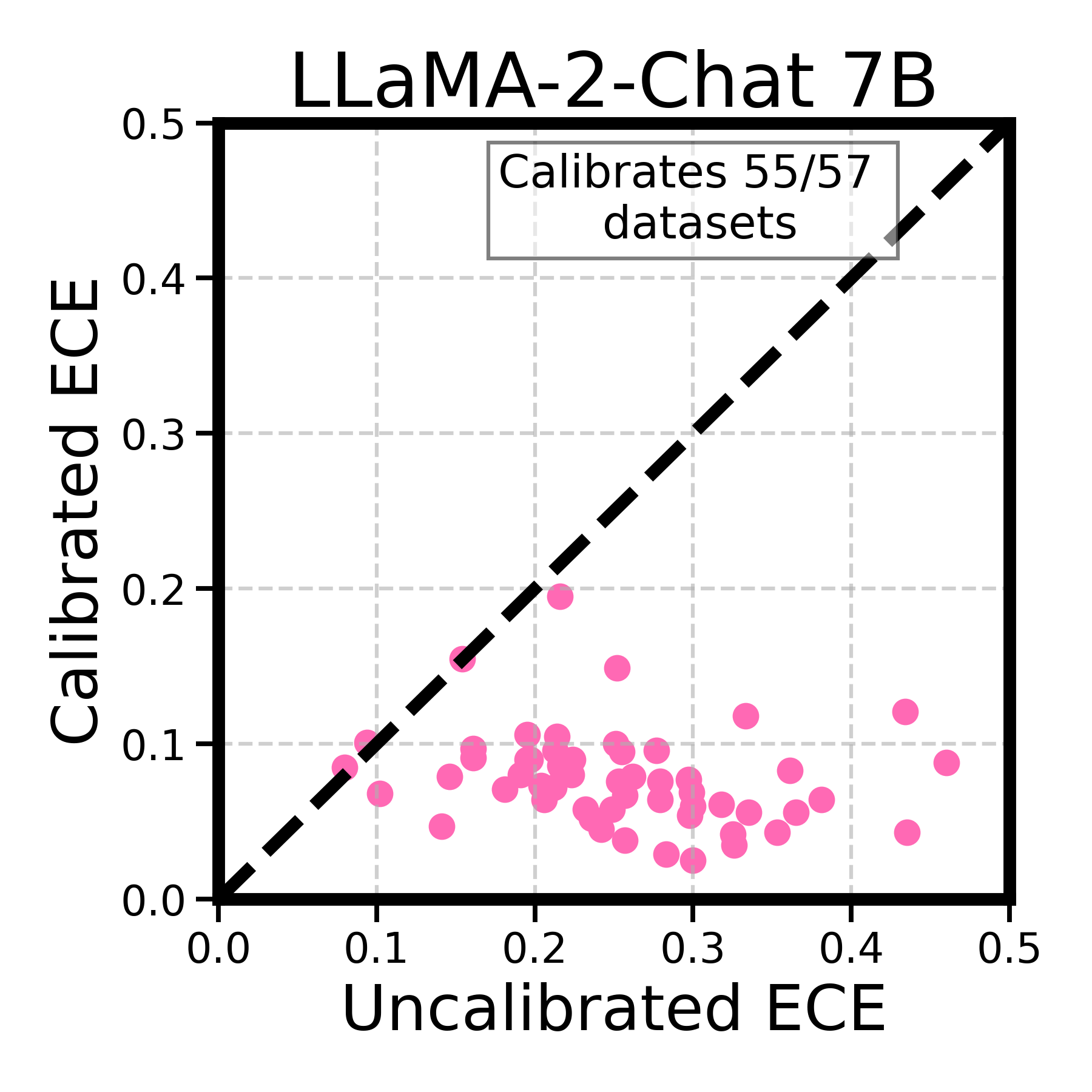} 
     \includegraphics[width=0.3\linewidth]{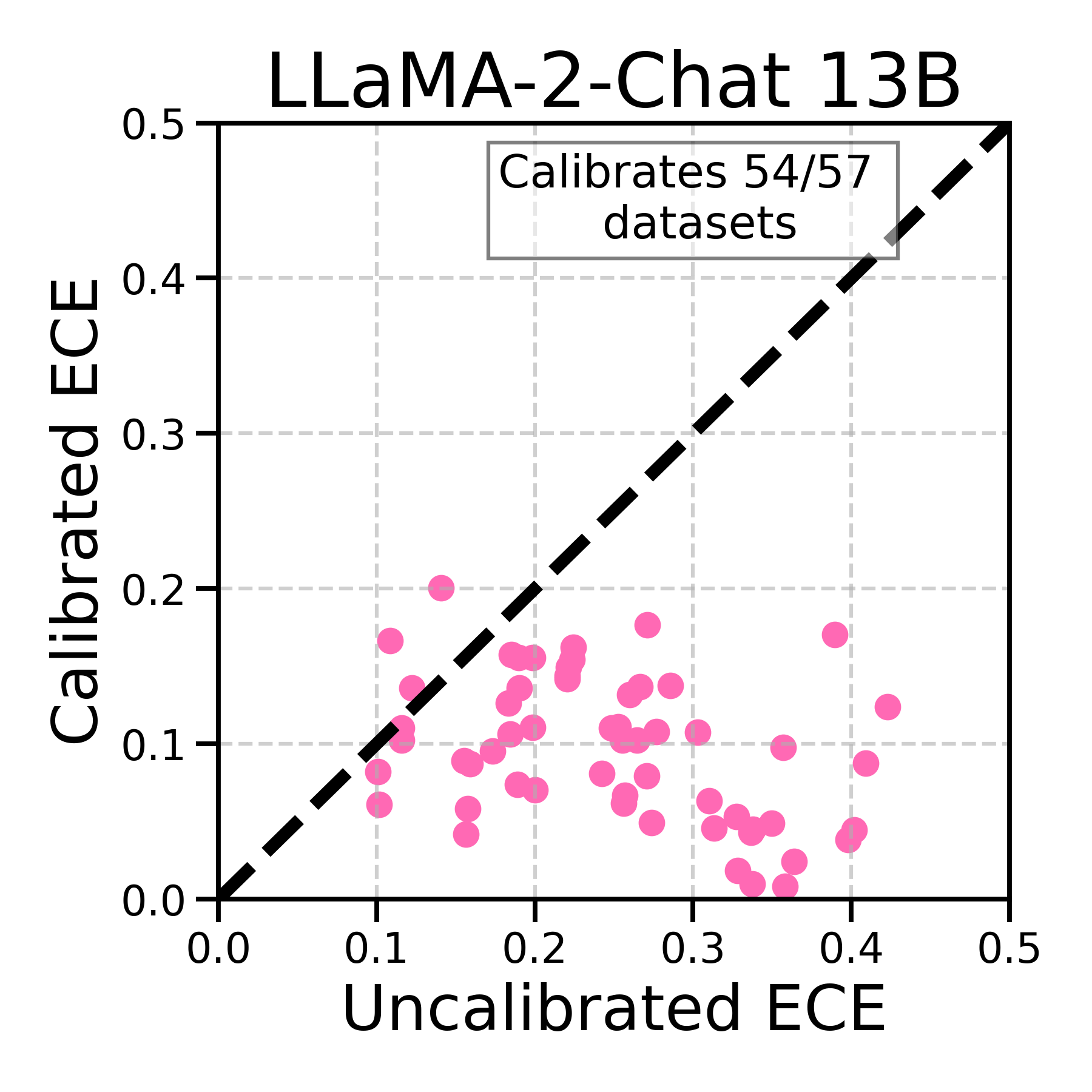} 
    \includegraphics[width=0.3\linewidth]{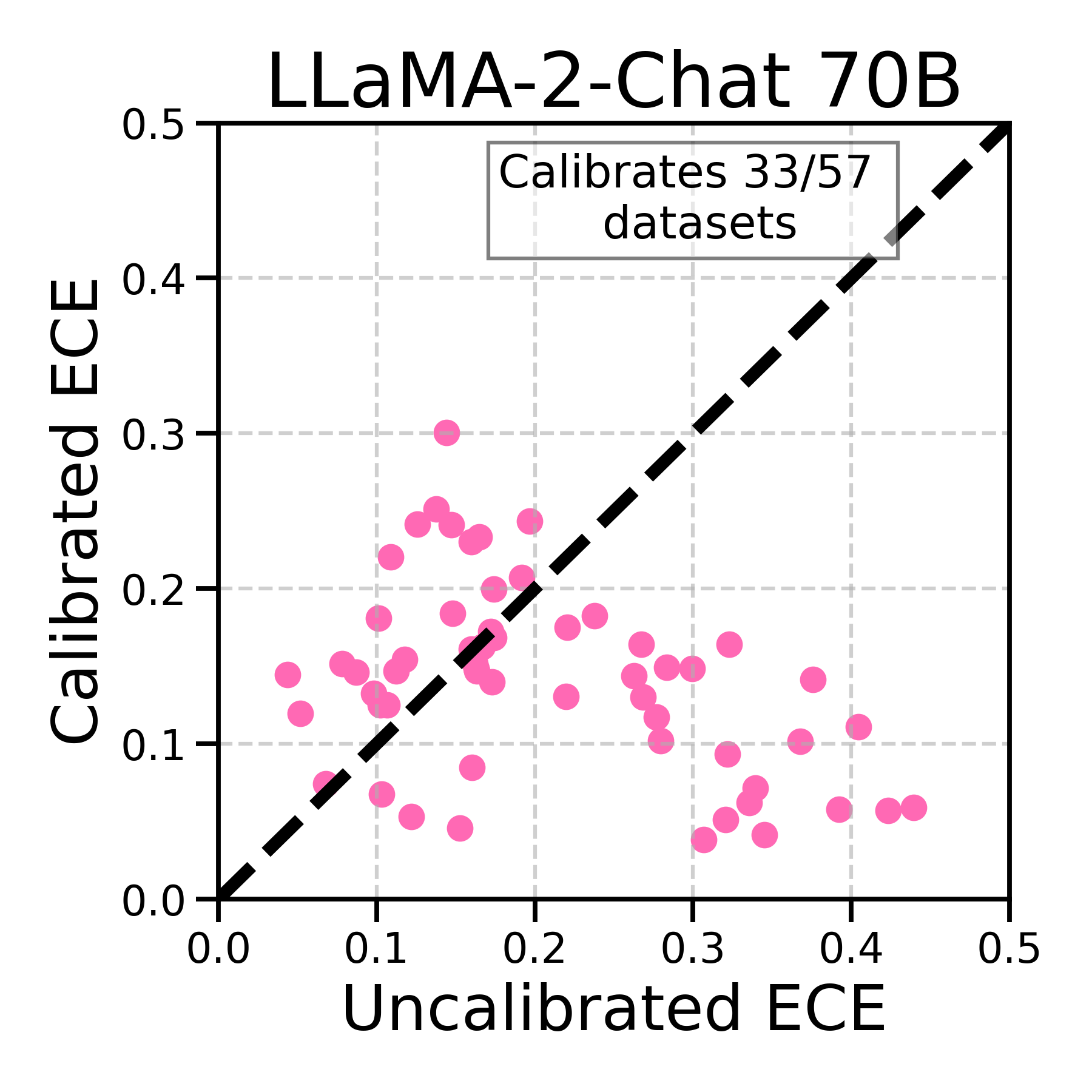}
\caption{{\textbf{$\modelname$ transfers across different model scales of LLaMA-2-Chat.} We use \llamaname~$\modelname$ predicted temperatures for calibrating \llamaname~(\emph{bold axes}),  LLaMA-2-Chat 13B and LLaMA-2-Chat 70B. In these plots, each dot represents a MMLU task. The x-coordinate is the ECE achieved by the uncalibrated model, the y-coordinate is the ECE achieved after calibrating the model with $\modelname$. We find that $\modelname$ predicted temperatures from the smaller models also improve calibration of larger models (shown in non-bold axes). }}
\label{figure:transfer-model}
\end{figure*}

\paragraph{Multiple-choice QA}
Our main calibration results for multiple-choice QA tasks are presented in two formats. We offer an overview of the overall calibration performance by calculating average results across all datasets. This includes 57 tasks from MMLU and 23 from BIG-bench. The average results and $95\%$ confidence intervals for MMLU are presented in \cref{table:MMLU-LLaMA}. We provide a finer granularity view through scatter plots illustrating performance on each task in \cref{figure:MMLU-ECE-scatter-LLaMA}. In this plots, points above the diagonal are instances where the calibration method fails to improve over the uncalibrated model, where $\modelname$ shows zero failure cases. 
Furthermore, we compare the predicted temperatures of $\modelname$, which uses no labeled data from the test task, with temperatures learned via temperature scaling that uses all the labeled data from the task in \cref{figure:MMLU-temperature}. We find $\modelname$'s predictions adapt to different tasks and align with learned temperatures.

The diverse nature of the BIG-bench tasks makes learning $\modelname$ more challenging. Nonetheless, the results for BIG-bench, included in \cref{app:bigbench}, $\modelname$ still outperforms other baselines with large margin, while stays within noise of TS-CV. Finally, we remark that the similar behavior of $\modelname$ holds for encoder-decoder model FLAN-T5-XL (see Appendix~\ref{app:T5}). $\modelname$ shows superior performance over other baseline methods. For the challenging BIG-bench, $\modelname$ with FLAN-T5-XL outperforms all competing approaches, producing lower ECE and TL-ECE scores than the strongest competitor, TS-CV, on $20/23$ tasks.

\paragraph{QA with Free Form Answers}
We use the MRQA shared task to evaluate $\modelname$'s efficacy for calibrating QA tasks with free form answers.  We train $\modelname$ with \llamaname~ on the MRQA training split and evaluate on the six non-overlapping datasets from the development split. The results are presented in \cref{table:mrqa-LLaMA}. Similar to multiple-choice QA tasks, we find that $\modelname$ substantially improves calibration of the uncalibrated model and performs favorably against competing methods.  Compared to the closest alternative, TS-CV, $\modelname$ produces lower ECE and TL-ECE scores on four of the six datasets. These results suggest broader applicability of $\modelname$ to (non-QA) free-form generation tasks.

\subsection{Analysis} \label{exp:analysis}

Here, we empirically scrutinize critical aspects of $\modelname$ using MMLU and \llamaname. Ablations analyzing the choice of $\modelname$ architecture, and batch sizes,  are in \cref{app:Ablation}.

\paragraph{$\modelname$ transfers across model scales and benchmarks}

\begin{figure}[ht]
  \centering
    \includegraphics[width=1.0\linewidth]{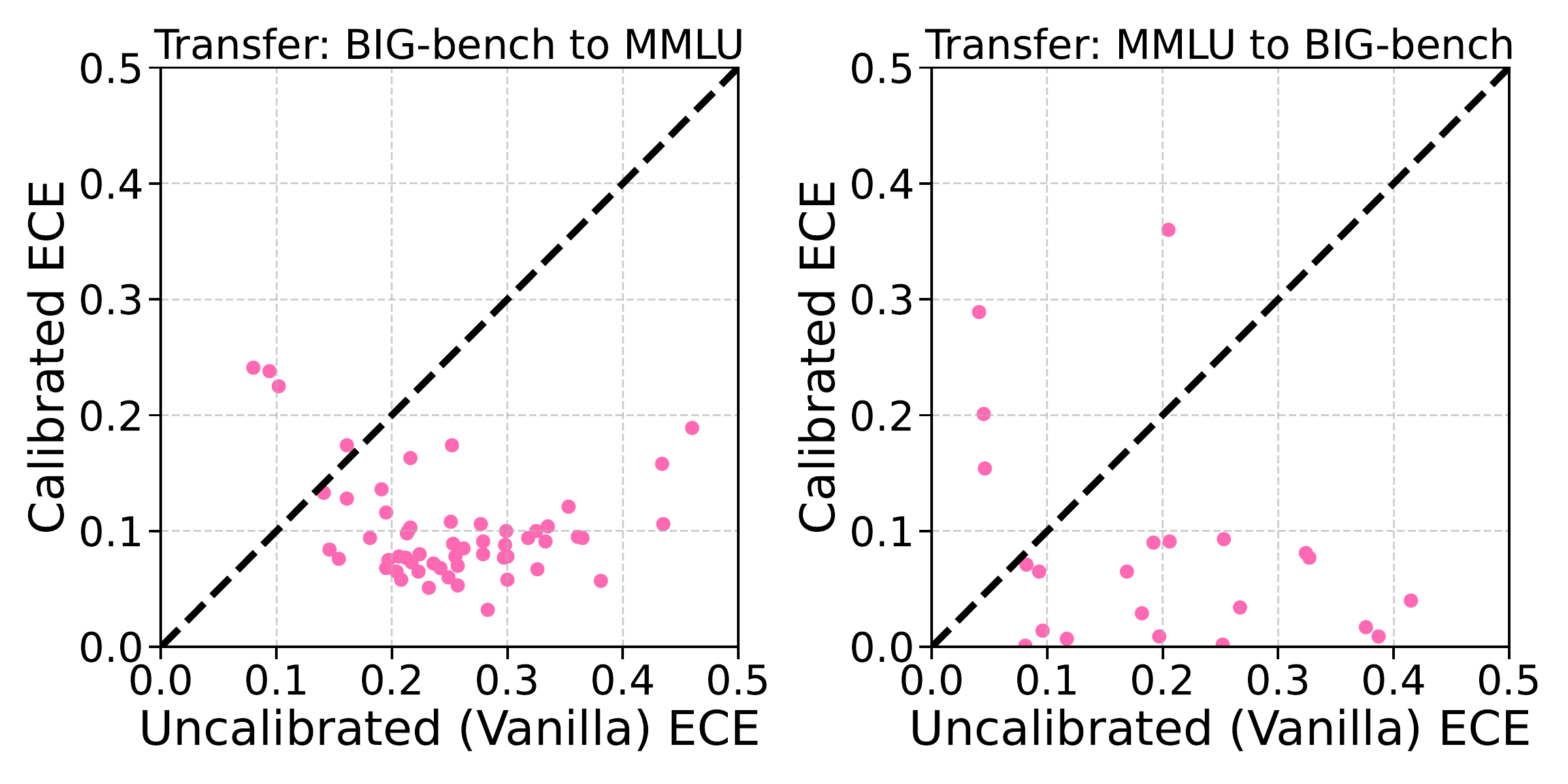}
    \vspace{-1.5em}
    \caption{{\textbf{$\modelname$ transfers across different datasets.} Applying \llamaname~$\modelname$ trained on BIG-bench calibrates MMLU and vice-versa.}}
    \label{figure:transfer-dataset}
    \vspace{-1em}
\end{figure}

We examine how well $\modelname$ can transfer its calibration capabilities across models of the same family but of different scales. The results in \cref{figure:transfer-model} demonstrate a significant finding, i.e., $\modelname$ trained with a smaller-scale model can effectively calibrate larger-scale models within the same family. This can substantially reduce the computational resources needed for calibrating large LLMs with $\modelname$.

We also explore the ability of $\modelname$ to transfer across different benchmark datasets. We train $\modelname$ on one benchmark dataset and test it on another. \cref{figure:transfer-dataset} shows that $\modelname$ effectively transfers across different benchmarks, highlighting the robustness of $\modelname$ to substantial data shifts.

\paragraph{$\modelname$ is effective in small labeled data regimes}
\begin{figure}[ht]
  \centering
    \includegraphics[width=0.8\linewidth]{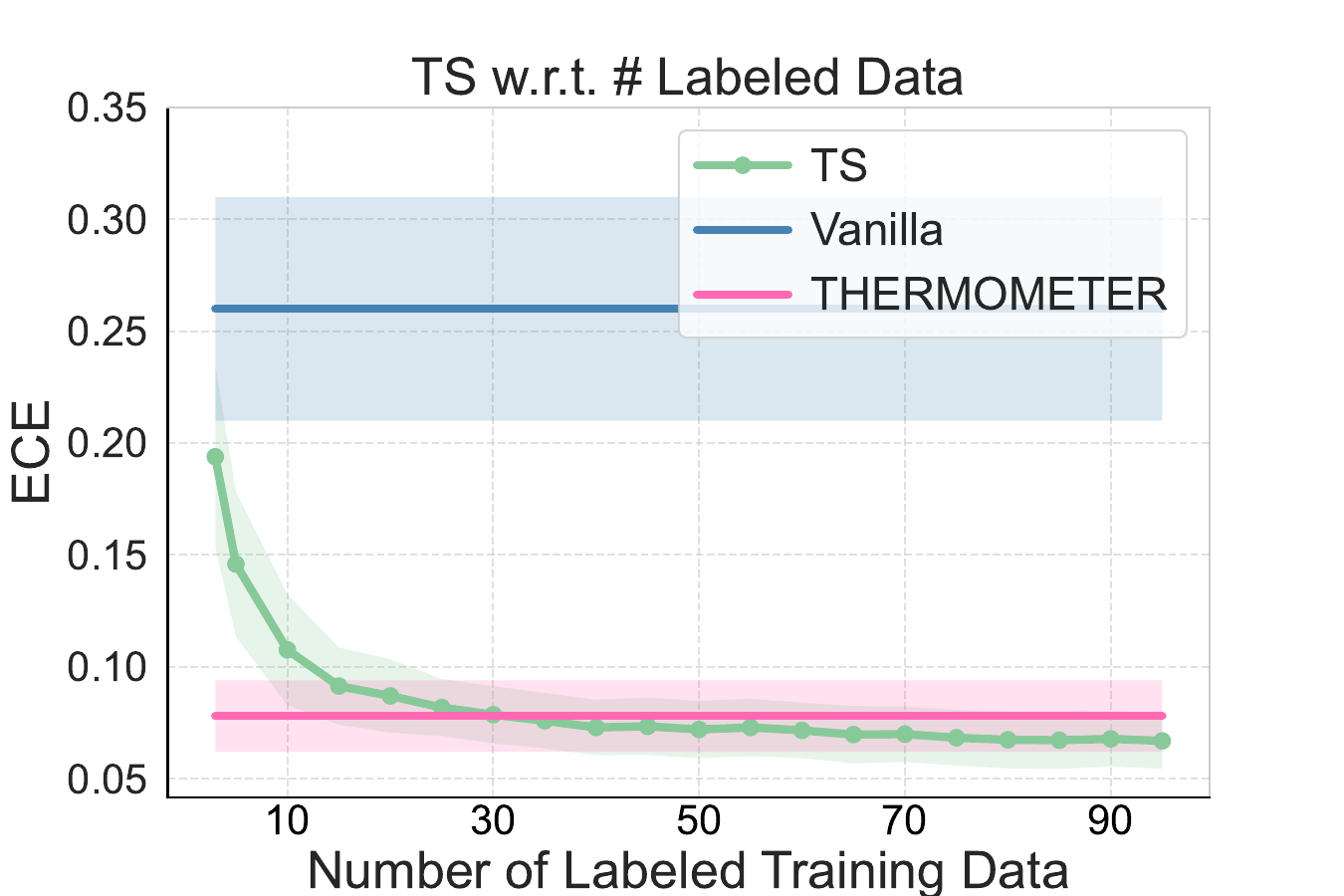}
    \vspace{-1em}
   \caption{\textbf{Temperature scaling vs. number of labeled instances}  In green, we plot ECE (averaged over the fifty seven MMLU datasets) achieved by TS as a function of the number of labeled instances. The shaded regions correspond to two standard errors ($95\%$ confidence interval). $\modelname$ outperforms TS when labeled data is less than thirty.}
    \label{figure:TS_num_data}
\end{figure}

$\modelname$ demonstrates a significant advantage in calibrating new test tasks by predicting the desired temperature using only unlabeled data. In contrast, TS requires labeled data for temperature tuning. The TS results in \cref{table:MMLU-LLaMA}, which utilize all available labeled data, serve as a lower-bound benchmark for our method. However, as shown in \cref{figure:TS_num_data}, $\modelname$ continues to outperform TS when a limited (but non-zero) amount of labeled data is available.  
This suggests that when labeled data for a task is scarce or nonexistent, it is more effective to rely on $\modelname$ than employing temperature scaling.

\paragraph{$\modelname$'s performance improves with number of training tasks}
\begin{figure}[ht]
  \centering
    \includegraphics[width=0.75\linewidth]{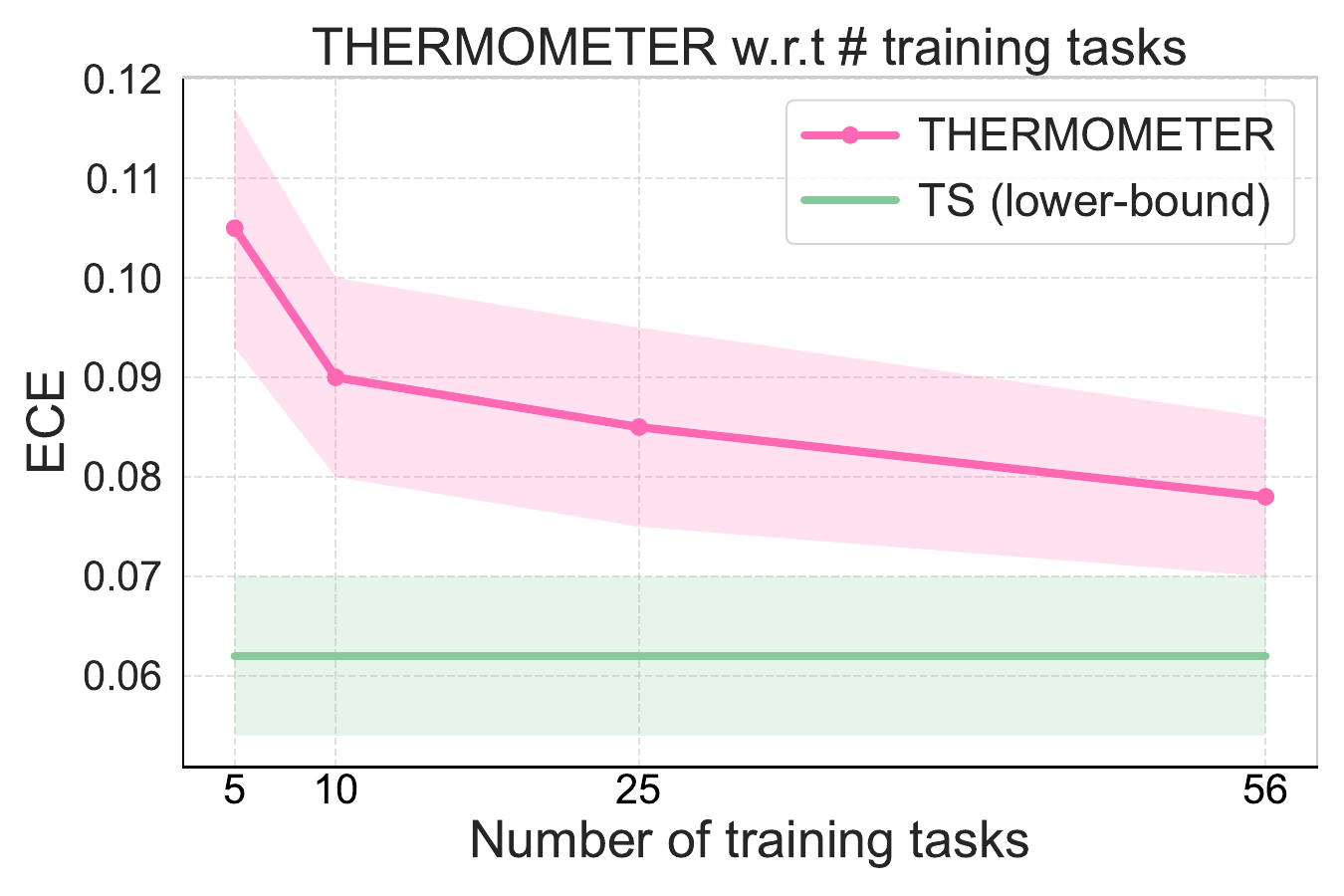}
    \vspace{-1em}
    \caption{\textbf{$\modelname$ performance vs. number of training tasks.}  $\modelname$'s calibration performance (average ECE over fifty seven MMLU tasks) improves as the number of training datasets increase. The shaded region represents two standard error.}
    \label{figure:Ablation_num_tasks}
\end{figure}
How many tasks do we need for training $\modelname$? We explore this by varying the number of training tasks. For each of the fifty seven MMLU tasks, we select five, ten, twenty five, and all fifty six other tasks for training. \cref{figure:Ablation_num_tasks} presents our results aggregated over all MMLU tasks. $\modelname$'s performance improves sharply between five and ten tasks and continues to improve linearly, albeit with a gentler slope, with more tasks.

\paragraph{$\modelname$'s aggregation procedure is more effective than sample-wise temperature} \label{app:sample_wise}

\begin{figure}[th]
    \centering
    \includegraphics[width=1.0\linewidth,clip]{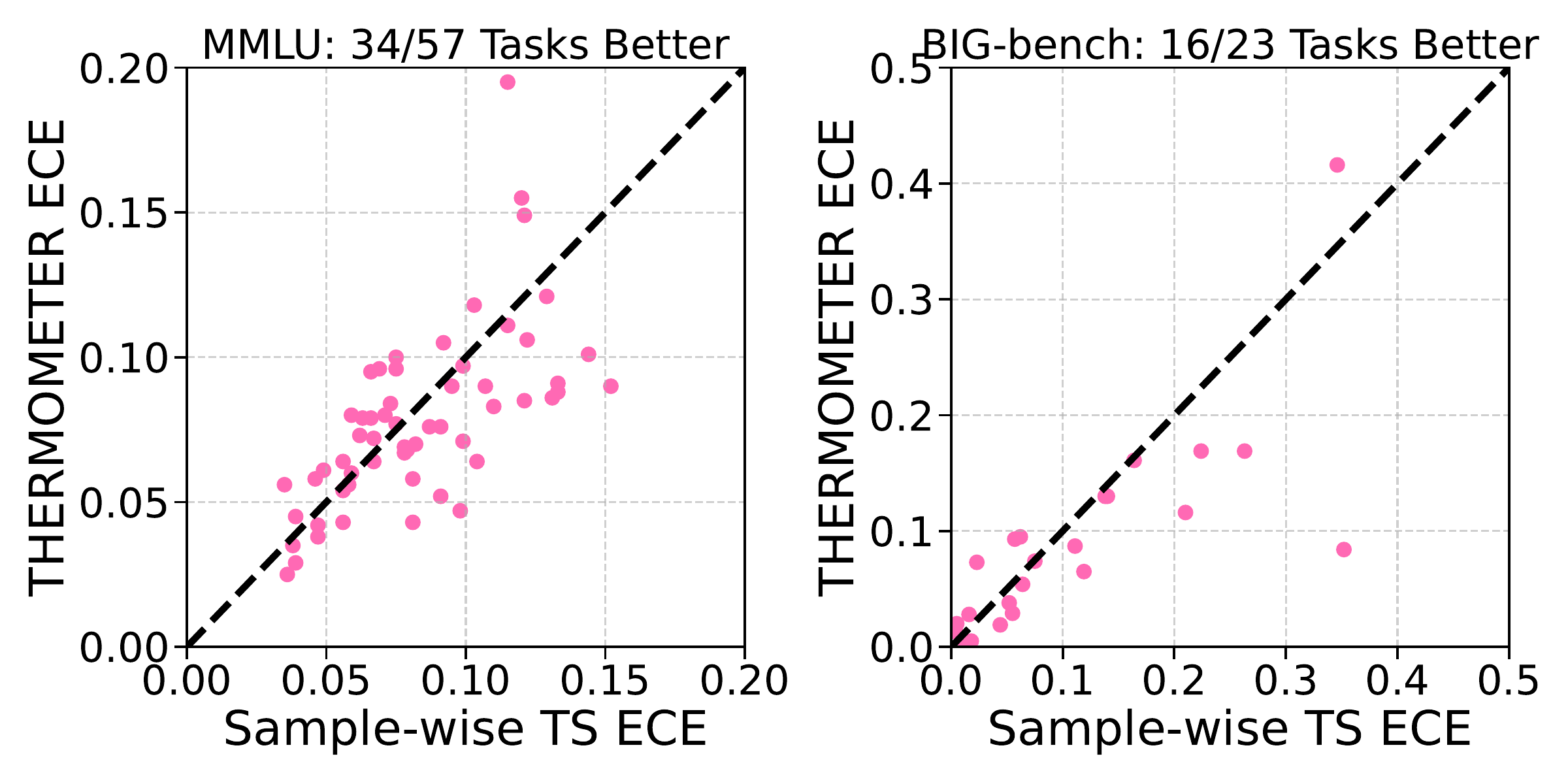}
    \vspace{-1.5em}
\caption{\textbf{Aggregation v.s. Sample-wise Temperature.} Sample-wise temperature scaling is less effective compared to $\modelname$.}
\label{figure:adaptive}
\end{figure}

We further evaluate the effectiveness of a variant of temperature scaling, i.e., sample-wise temperature scaling approach~\citep{joy2023sample} that learns a unique temperature for each data point. While their single-task approach doesn't apply here, we can adapt $\modelname$ to predict temperatures on a sample-wise basis, rather than using a single temperature for all samples in a dataset. The results presented in \cref{figure:adaptive} show that while sample-wise temperature scaling is effective to a degree, it is sub-optimal when compared to aggregation across data samples. Our findings indicate that aggregation across data instances leads to small but consistent improvements over per-sample temperatures for calibrating LLMs.  

\section{Concluding Remarks}    \label{sec: conclusion}
This work introduces $\modelname$, a simple yet effective approach for calibrating LLMs. Through comprehensive empirical evaluations we find that it improves calibration, is robust to data shifts, and is computationally efficient. While we demonstrate the effectiveness of $\modelname$ in question answering tasks with free-form answers, the potential for its application extends far beyond. Future directions include adapting $\modelname$ for other complex free-form generation tasks, such as summarization and translation, and applying $\modelname$ to larger LLMs.

\flushcolsend
\clearpage
\section*{Impact Statement}
This paper presents work whose goal is to advance the field
of Machine Learning. There are many potential societal
consequences of our work, none of which we feel must be
specifically highlighted here.

\section*{Acknowledgements}
We are grateful to Natalia Martinez Gil, Dan Gutfreund, Farzaneh Mirzazadeh, and Veronika Thost for their feedback
on the manuscript and to Inkit Padhi for help with compute infrastructure

\bibliography{refs}
\bibliographystyle{icml2024}

\flushcolsend
\clearpage
\appendix
\onecolumn
\section{Additional Results} \label{app:results}

\subsection{$\modelname$ Consistently Performs Well on BIG-bench} \label{app:bigbench}
\begin{figure*}[!t]
\vspace{-0.5em}
    \centering
    \includegraphics[width=0.75\linewidth,clip]{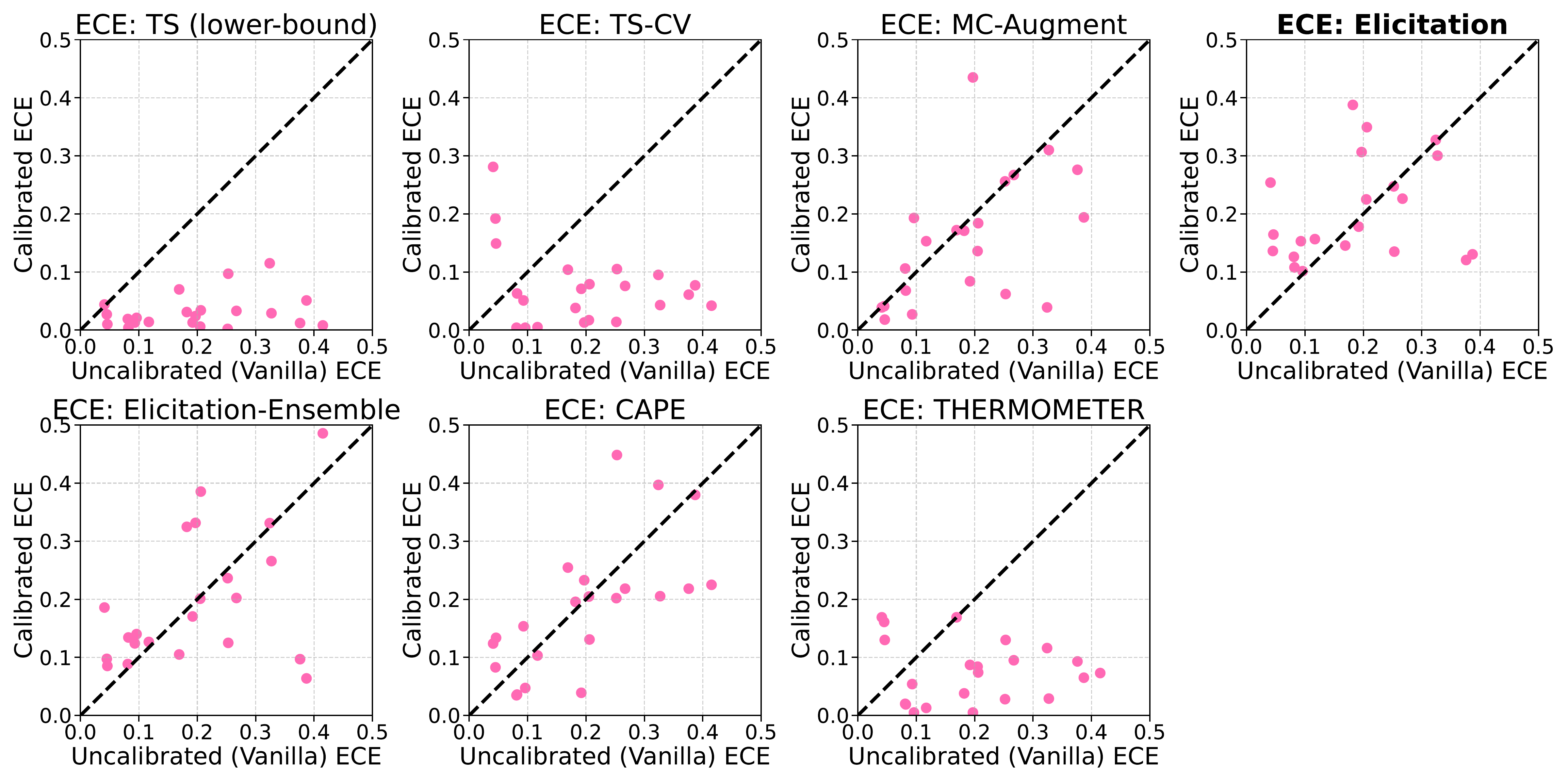}
\caption{\textbf{\llamaname~Scatter Plots: ECE Score of 23 BIG-bench Datasets.} The x-axis and y-axis represent ECE score of uncalibared and calibrated model, respectively. $\modelname$ largely reduces calibration error, and rarely fails to improve calibration.}
\label{figure:BIG-bench-ECE-scatter-LLaMA}
\end{figure*}

\begin{figure*}[!t]
    \centering
    \includegraphics[width=0.3\linewidth,clip]{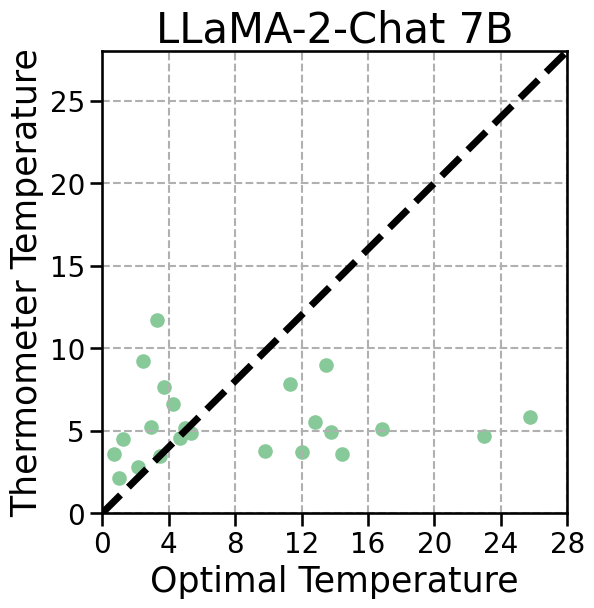}
\caption{\textbf{$\modelname$ against temperature scaling on BIG-bench (\llamaname).} Comparison of $\modelname$ predicted temperature and optimal temperature obtained by temperature scaling. While there are a few outliers, $\modelname$ predicted temperatures are still correlated with the optimal temperatures.}
\vspace{-1em}
\label{figure:BIG-bench-temperature-llama}
\end{figure*}

The average calibration performance of $\modelname$ on the BIG-bench dataset is presented in Table~\ref{table:BIG-bench-LLaMA}. Additionally, scatter plots illustrating these results are detailed in Figure~\ref{figure:BIG-bench-ECE-scatter-LLaMA}. The comparison between $\modelname$'s predicted temperatures and the optimal temperatures obtained through temperature scaling is depicted in Figure~\ref{figure:BIG-bench-temperature-llama}. Despite the diversity and complexity of the BIG-bench datasets, which pose a challenge in predicting the desired temperature, $\modelname$ still maintains a superior performance compared to other baseline methods.

\begin{table*}[!t]
  \begin{center}
  \footnotesize
  \captionsetup{font=small}
  \caption{\textbf{\llamaname~Average Calibration Performance on BIG-bench.} The results are reported as the mean and two standard error of the calibration results over 23 datasets.}
   \resizebox{0.65\columnwidth}{!}{
  \begin{tabular}{cccccccc}
    \hline
    \textbf{Methods}  & \textbf{ECE} &  \textbf{TL-ECE}  & \textbf{MCE} & \textbf{NLL} & \textbf{Brier}\\
    \hline
    TS (lower-bound) & 0.044$\pm$0.014 & 0.089$\pm$0.025 & 0.106$\pm$0.027 & 1.229$\pm$0.130 & 0.189$\pm$0.013 \\
    \hdashline
    Vanilla & 0.233$\pm$0.035 & 0.263$\pm$0.037 & 0.467$\pm$0.055 & 1.539$\pm$0.171 & 0.232$\pm$0.021 \\
    TS-CV & \textbf{0.087$\pm$0.020} & \textbf{0.118$\pm$0.027} & \textbf{0.240$\pm$0.041} & \textbf{1.258$\pm$0.129} & \textbf{0.195$\pm$0.013} \\
    MC-Augment &0.186$\pm$0.031 & 0.231$\pm$0.033 & 0.436$\pm$0.043 & 1.466$\pm$0.180 & 0.214$\pm$0.016 \\
    Elicitation &0.225$\pm$0.048 &0.263$\pm$0.059 &0.745$\pm$0.085 & $\slash$ &$\slash$ \\
    Elicitation-Ensemble & 0.202$\pm$0.049 &0.241$\pm$0.063	&0.753$\pm$0.068  & $\slash$ & $\slash$\\
    CAPE &0.214$\pm$0.073 &0.235$\pm$0.076 &0.475$\pm$0.107 & 1.567$\pm$0.365 & 0.231$\pm$0.043 \\
    \rowcolor{lightgray}
    \textbf{Thermometer} & \textbf{0.090$\pm$0.018} & \textbf{0.125$\pm$0.026} & \textbf{0.243$\pm$0.038} & \textbf{1.261$\pm$0.132} & \textbf{0.195$\pm$0.013} \\
    \hline
  \end{tabular}}
  \label{table:BIG-bench-LLaMA}
  \end{center}
\end{table*}

\clearpage
\subsection{$\modelname$ Consistently Performs Well on Encoder-decoder LLMs} \label{app:T5}
To demonstrate $\modelname$'s robustness with respect to different types of language models, we also conduct a set of additional experiments on encoder-decoder model FLAN-T5-XL with three billion parameters. We empirically observe that elicitation-based calibration methods~\citep{tian2023just, xiong2023can} are not even functional on FLAN-T5-XL model thus omit reporting their results, i.e., by providing the crafted prompts, FLAN-T5-XL often fails to generate the corresponding verbalized confidence. This indicates that elicitation-based methods~\citep{tian2023just, xiong2023can} that rely on the ability of a model to faithfully follow nuanced instructions generally do not perform well for smaller and mid-sized open source models. In fact, the authors in~\citep{tian2023just} concede that even for Llama2-70b, ``The verbal calibration of the open source model Llama-2-70b-chat is generally weaker than that of closed source models''.

\begin{figure*}[!t]
    \centering
    \includegraphics[width=0.65\linewidth,clip]{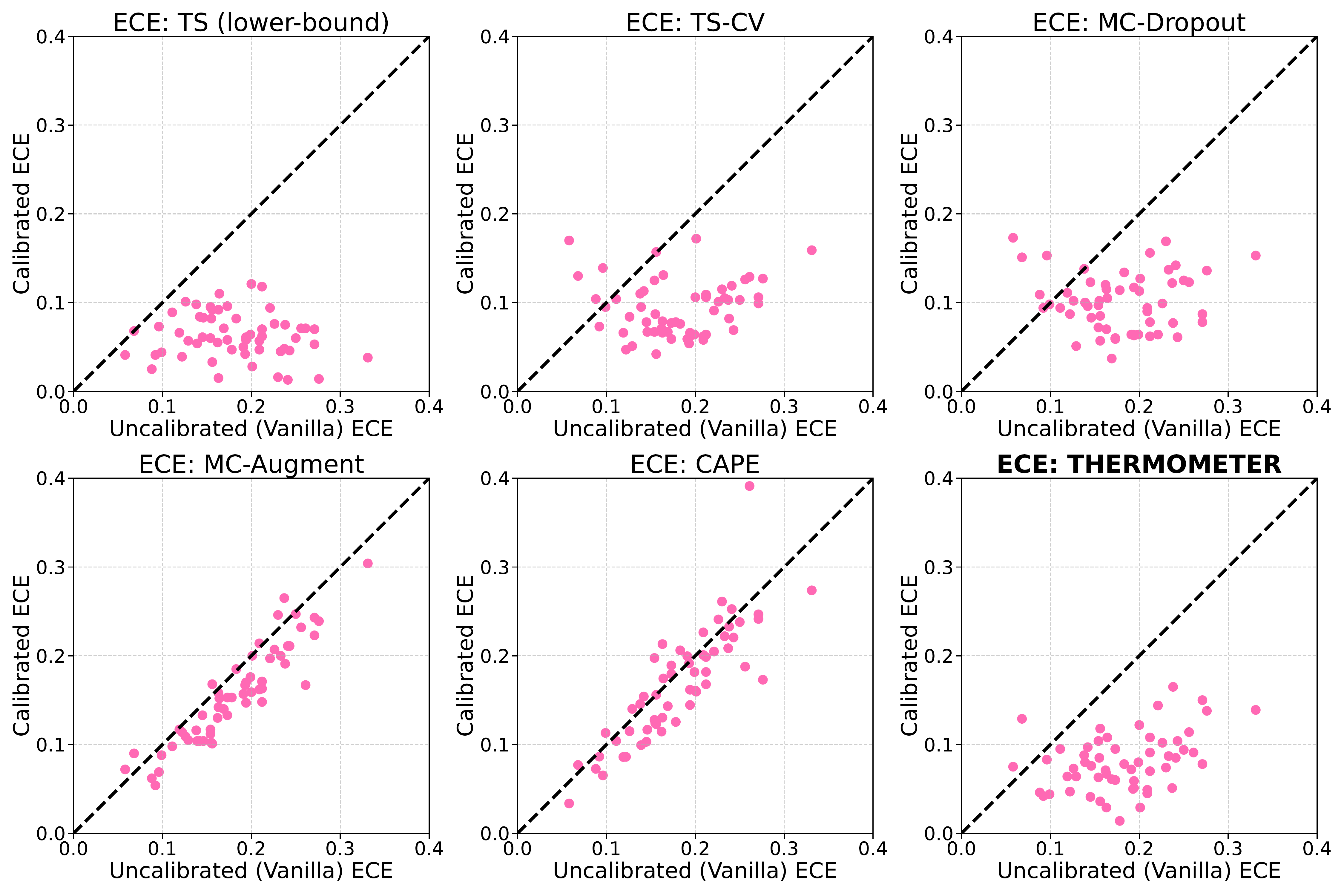}
\caption{\small\textbf{FLAN-T5-XL Scatter Plots: ECE Score of 57 MMLU Datasets.} The x-axis and y-axis represent ECE score of uncalibared and calibrated model, respectively. $\modelname$ largely reduces calibration error, and rarely fails to improve calibration. }
\label{figure:MMLU-ECE-scatter-T5}
\vspace{-1em}
\end{figure*}

\begin{figure*}[!t]
\vspace{-1em}
    \centering
    \includegraphics[width=0.65\linewidth,clip]{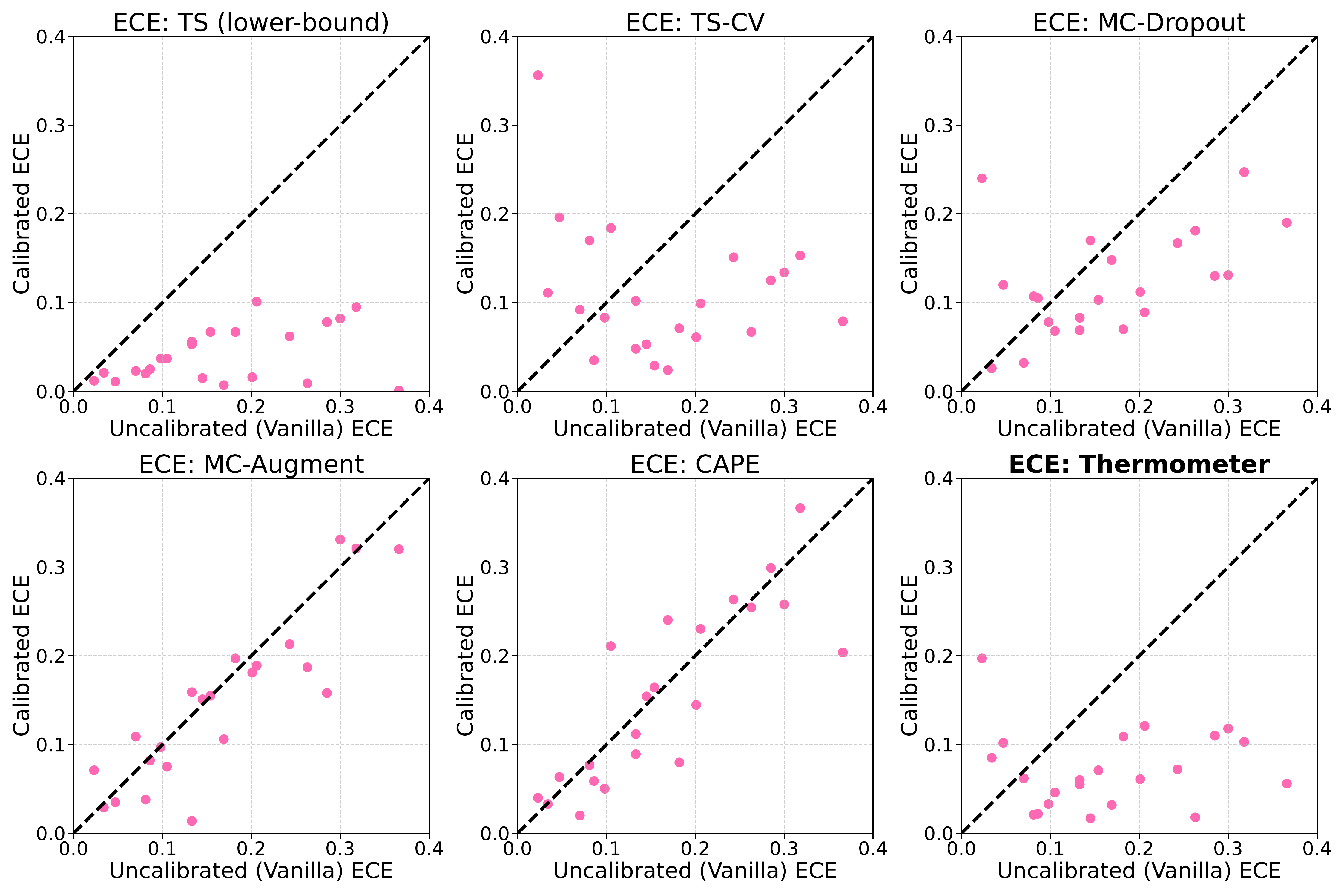}
\caption{\textbf{FLAN-T5-XL Scatter Plots: ECE Score of 23 BIG-bench Datasets.} The x-axis and y-axis represent ECE score of uncalibared and calibrated model, respectively. $\modelname$ largely reduces calibration error, and rarely fails to improve calibration. }
\label{figure:BIG-bench-ECE-scatter-T5}
\end{figure*}

\begin{figure}[!t]
    \centering
    \begin{tabular}{c|c}
     \includegraphics[width=0.3\linewidth,clip]{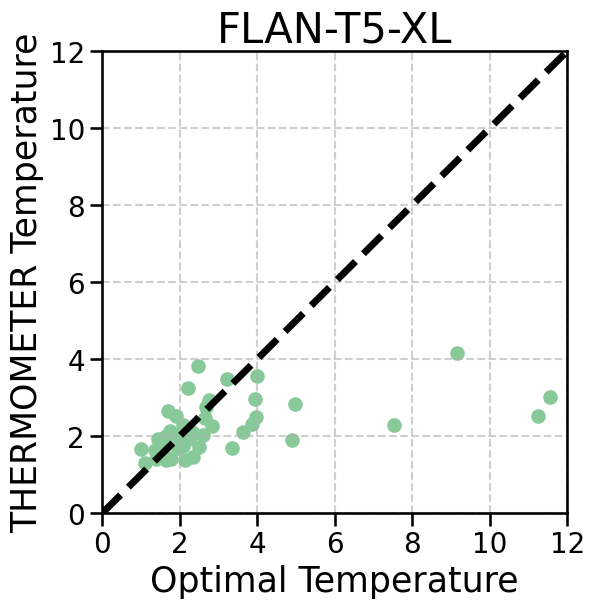}&
    \includegraphics[width=0.3\linewidth,clip]{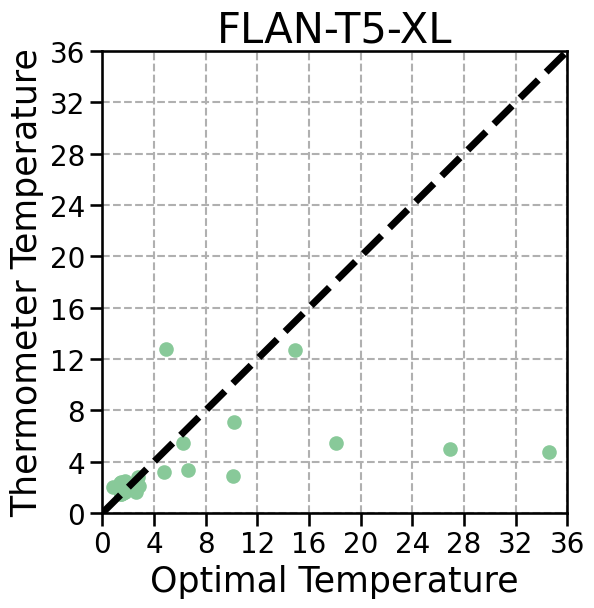}
    \end{tabular}
\caption{\textbf{$\modelname$ against temperature scaling on MMLU and BIG-bench (FLAN-T5-XL).} Comparison of $\modelname$ predicted temperature and optimal temperature obtained by temperature scaling. While there are a few outliers, $\modelname$ predicted temperatures are still correlated with the optimal temperatures.}
\label{figure:temperature-T5}
\end{figure}

We evaluate the calibration performance of different methods on both MMLU and BIG-bench. The average calibration performance of $\modelname$ on MMLU datasets and BIG-bench datasets are shown in Table~\ref{table:MMLU-T5} and Table~\ref{table:BIG-bench-T5}, respectively. We also provide the scatter plots of comparison for MMLU and BIG-bench in Figure~\ref{figure:MMLU-ECE-scatter-T5} and Figure~\ref{figure:BIG-bench-ECE-scatter-T5}, respectively. Finally, The comparison between $\modelname$'s predicted temperatures and the optimal temperatures obtained through temperature scaling is presented in Figure~\ref{figure:temperature-T5}. Similar to decode-only model \llamaname, $\modelname$ consistently outperforms other baselines. Furthermore, we further validate that $\modelname$ also transfers well across different model scales of Flan-T5 in Figure~\ref{fig:transfer-ts-cv}.

\begin{table*}[!t]
  \begin{center}
  \footnotesize
  \caption{\textbf{FLAN-T5-XL Average Calibration Performance on MMLU.} The results are reported as the mean and two standard error of the calibration results over 57 datasets. TS serves as the lower-bound as it has access to the labeled data of testing task.}
  \resizebox{0.65\columnwidth}{!}{
  \begin{tabular}{cccccccc}
    \hline
    \textbf{Methods}  & \textbf{ECE} &  \textbf{TL-ECE}  & \textbf{MCE} & \textbf{NLL} & \textbf{Brier} \\
    \hline
    TS (lower-bound) & 0.063$\pm$0.006 & 0.128$\pm$0.010 & 0.249$\pm$0.037 & 1.141$\pm$0.054 & 0.153$\pm$0.008 \\
    \hdashline
    Vanilla & 0.181$\pm$0.013 & 0.215$\pm$0.014 & 0.448$\pm$0.049 & 1.286$\pm$0.074 & 0.167$\pm$0.010 \\
    TS-CV & 0.093$\pm$0.007 & 0.146$\pm$0.010 & 0.314$\pm$0.053 & 1.160$\pm$0.054 & 0.155$\pm$0.009  \\
    MC-Dropout & 0.107$\pm$0.015 & 0.159$\pm$0.012 & 0.372$\pm$0.041 & 1.301$\pm$0.065 & 0.171$\pm$0.007 \\
    MC-Augment & 0.156$\pm$0.015 & 0.198$\pm$0.014 & 0.431$\pm$0.045 & 1.266$\pm$0.067 & 0.167$\pm$0.008 \\
    CAPE &  0.167$\pm$0.017 & 0.210$\pm$0.015 & 0.410$\pm$0.048 &  1.271$\pm$0.072 & 0.166$\pm$ 0.009 \\
    \rowcolor{lightgray}
    \textbf{\modelname} & \textbf{0.080$\pm$0.008} & \textbf{0.139$\pm$0.011} & \textbf{0.319$\pm$0.042} & \textbf{1.154$\pm$0.055} & \textbf{0.155$\pm$0.009} \\
    \hline
  \end{tabular}}
  \label{table:MMLU-T5}
  \end{center}
\end{table*}

\begin{table*}[!t]
  \begin{center}
  \footnotesize
  \captionsetup{font=small}
  \caption{\textbf{FLAN-T5-XL Average Calibration Performance on BIG-bench.} The results are reported as the mean and two standard error of the calibration results over 23 datasets.}
  \resizebox{0.65\columnwidth}{!}{
  \begin{tabular}{cccccccc}
    \hline
    \textbf{Methods}  & \textbf{ECE} &  \textbf{TL-ECE}  & \textbf{MCE} & \textbf{NLL} & \textbf{Brier} \\
    \hline
    TS (lower-bound) & 0.043$\pm$0.006 & 0.109$\pm$0.022 & 0.179$\pm$0.042 & 1.087$\pm$0.147 & 0.157$\pm$0.013 \\
    \hdashline
    Vanilla & 0.192$\pm$0.032 & 0.230$\pm$0.033 & 0.444$\pm$0.055 & 1.417$\pm$0.235 & 0.183$\pm$0.017 \\
    TS-CV & 0.121$\pm$0.018 & 0.175$\pm$0.026 & 0.306$\pm$0.039 & 1.170$\pm$0.150 & 0.168$\pm$0.013 \\
    MC-Dropout & 0.130$\pm$0.015 & 0.184$\pm$0.023 & 0.360$\pm$0.056 & 1.311$\pm$0.184 & 0.180$\pm$0.014  \\
    MC-Augment & 0.167$\pm$0.028 & 0.211$\pm$0.030 & 0.374$\pm$0.058 & 1.378$\pm$0.213 & 0.186$\pm$0.017 \\
    CAPE &  0.176$\pm$0.057 & 0.213$\pm$0.061 & 0.432$\pm$0.122 &  1.387$\pm$0.242 &  0.181$\pm$ 0.033 \\
    \rowcolor{lightgray}
    \textbf{Thermometer} & \textbf{0.078$\pm$0.011} & \textbf{0.138$\pm$0.021} & \textbf{0.220$\pm$0.035} & \textbf{1.113$\pm$0.148} & \textbf{0.160$\pm$0.013} \\
    \hline
  \end{tabular}}
  \label{table:BIG-bench-T5}
  \end{center}
\end{table*}

\begin{figure*}[!t]
    \centering
    \includegraphics[width=0.25\linewidth]{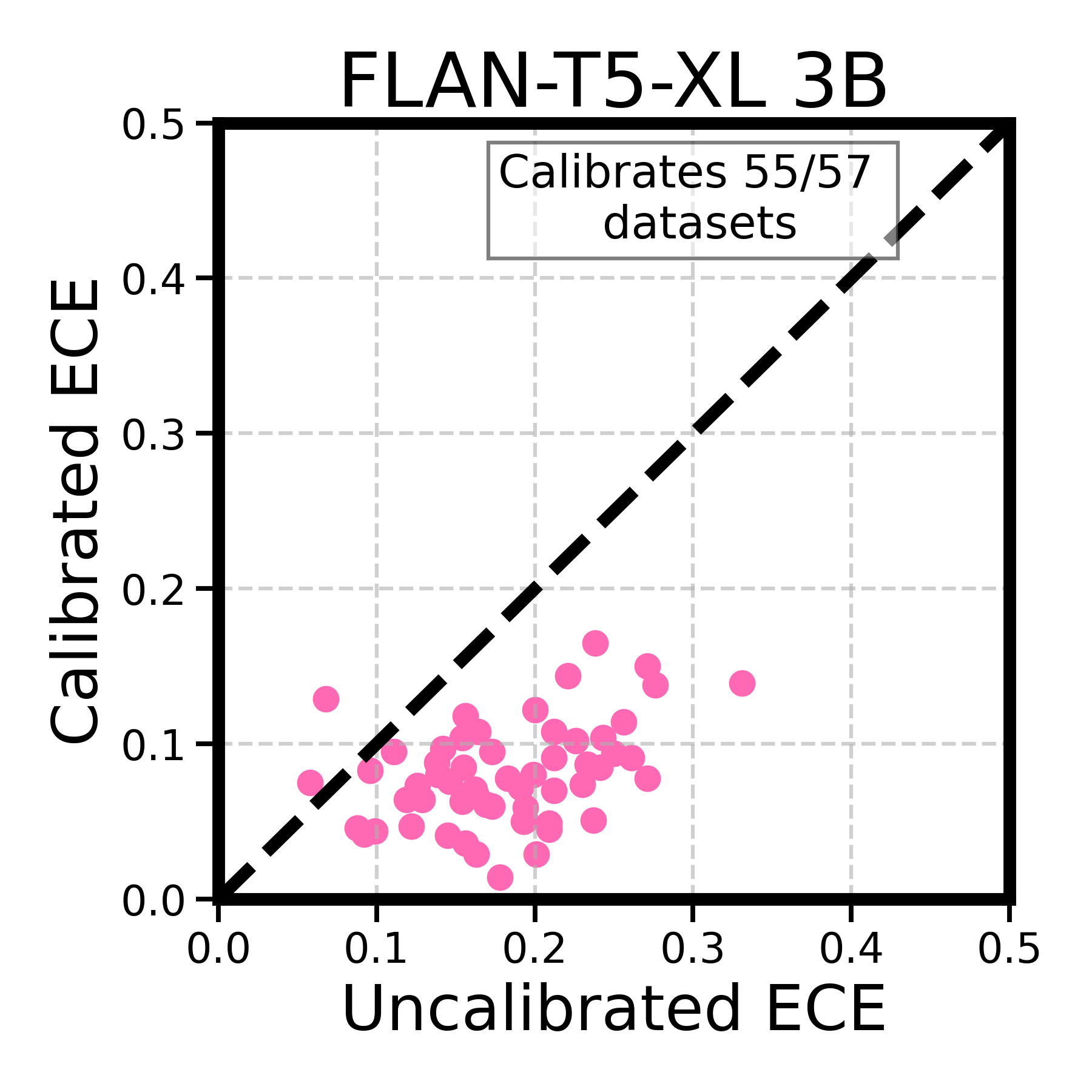} 
     \includegraphics[width=0.25\linewidth]{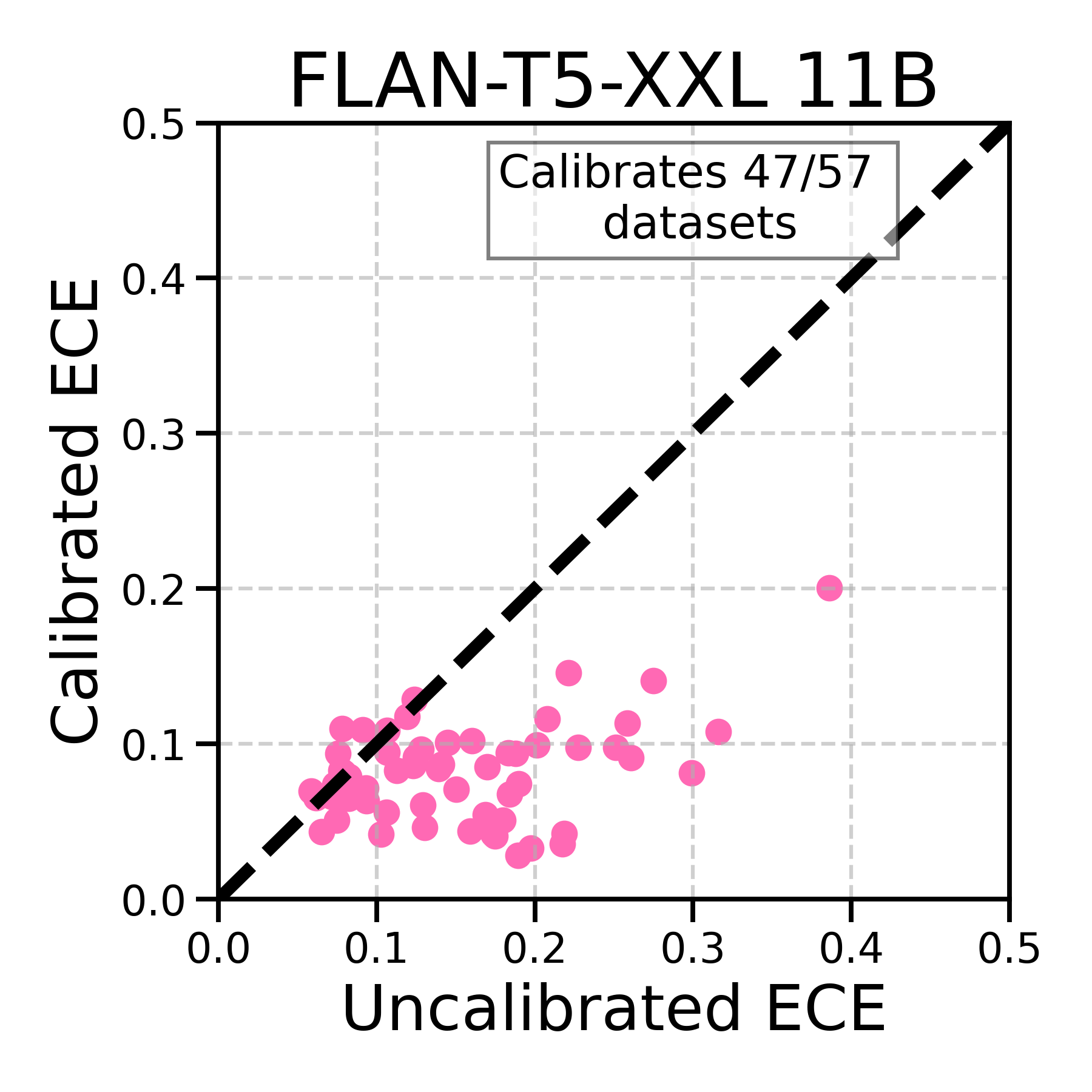} 
    \includegraphics[width=0.25\linewidth]{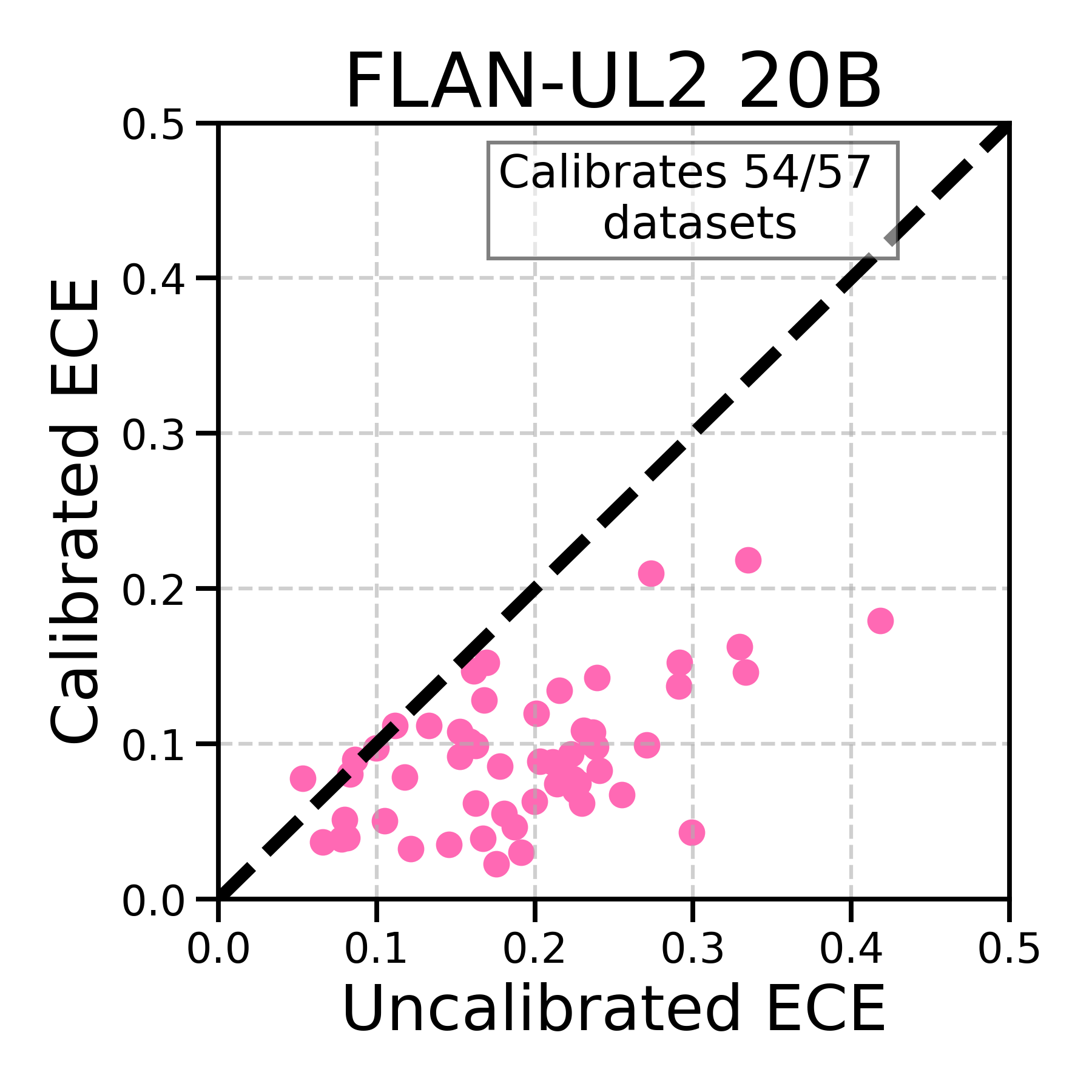}
\caption{{\textbf{$\modelname$ transfers across different model scales of FLan-T5.} We use FLan-T5-XL 3B $\modelname$ predicted temperatures to calibrate FLan-T5-XL 3B (\emph{bold axes}),  FLan-T5-XXL 11B and Flan-UL2 20B. In these plots, each dot represents a MMLU task. The x-coordinate is the ECE achieved by the uncalibrated model, the y-coordinate is the ECE achieved after calibrating the model with $\modelname$. We find that $\modelname$ predicted temperatures from the smaller models also improve calibration of larger models (shown in non-bold axes). }}
\label{figure:transfer-model-T5}
\end{figure*}

\newpage
\subsection{$\modelname$ Shows Stronger Transfer-ability Than TS-CV}
\begin{figure}[!t]
    \centering
    \includegraphics[width=0.25\linewidth]{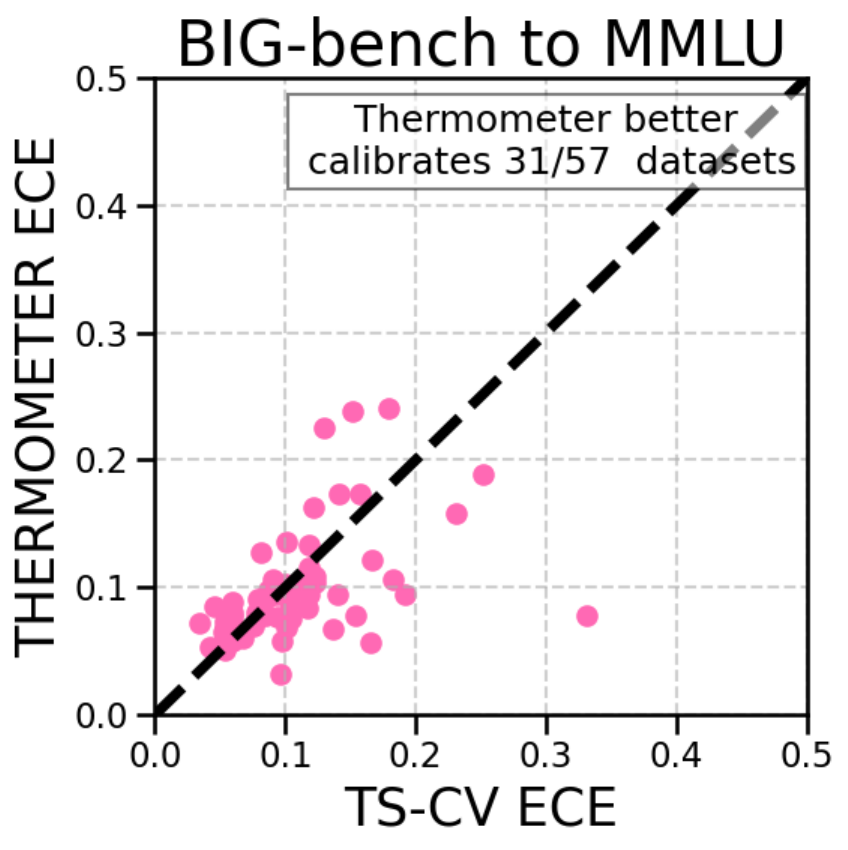} 
    \includegraphics[width=0.25\linewidth]{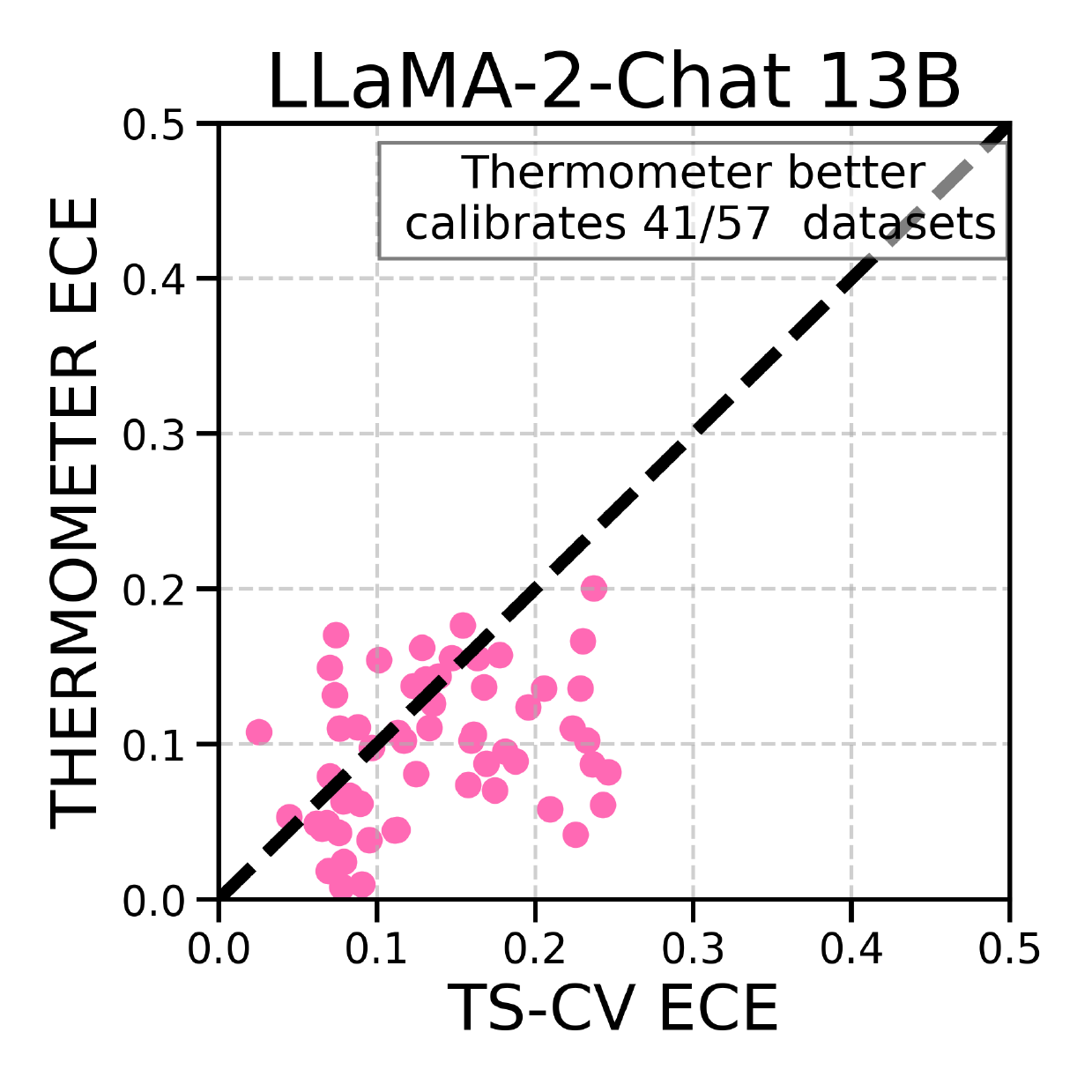}\\
     \includegraphics[width=0.25\linewidth]{Figures/TS-CV_transfer_mmlu.pdf} 
    \includegraphics[width=0.25\linewidth]{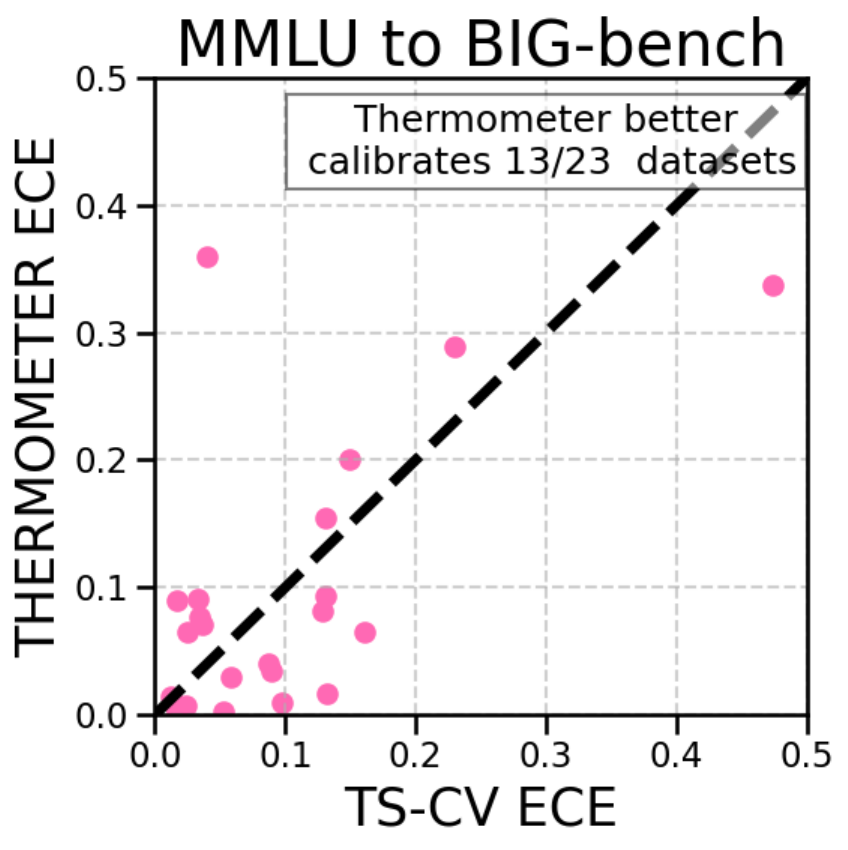}\\
\caption{\small{\textbf{Compare Transfer-ability of $\modelname$ with TS-CV.} \emph{Top}: We use FLan-T5-XL 3B $\modelname$ predicted temperatures to calibrate 20B Flan-UL2 in the left subplot, and \llamaname~$\modelname$ predicted temperatures for calibrating LLaMA-2-Chat 13B in the right subplot. \emph{Bottom}: Applying \llamaname~$\modelname$ trained on BIG-bench calibrates MMLU and vice-versa.}}
\label{fig:transfer-ts-cv}
\end{figure}

In Figure~\ref{figure:transfer-model} and Figure~\ref{figure:transfer-dataset}, we demonstrate the strong transfer-ability of $\modelname$ across different model scales and datasets. This is the another advantage of $\modelname$ over those inference-based methods~\citep{wei-zou-2019-eda, tian2023just, xiong2023can, jiang2023calibrating}, which do have the capability to transfer. TS-CV is the only baseline that has transfer-ability, so we also conduct an additional experiment to compare the transfer-ability of $\modelname$ with the TS-CV baseline. As illustrated Figure~\ref{fig:transfer-ts-cv}, Thermometer typically shows stronger transfer-ability than TS-CV, producing lower calibration errors. 


\clearpage
\subsection{Ablation Studies} \label{app:Ablation}
To obtain a deeper insight into the factors influencing $\modelname$ performance, we conduct an ablation study focusing on various elements such as the model architecture, the value of the regularizer weight, training batch size, and the size of the test data used during inference. The results are detailed in Appendix~\ref{app:Ablation}.
\begin{table*}[!t]
  \begin{center}
  \footnotesize
  \captionsetup{font=small}
  \caption{\textbf{Ablation Study of $\modelname$ Architecture.} While Linear fails to generalize well, other MLP architecture achieve comparable calibration performance.}
  \begin{tabular}{cccccccc}
    \hline
    \textbf{Architecture}  & \textbf{ECE} &  \textbf{TL-ECE}  & \textbf{MCE} & \textbf{NLL} & \textbf{Brier}\\
    \hline
    Linear & 0.102$\pm$0.006 & 0.151$\pm$0.007 & 0.322$\pm$0.015 & 1.241$\pm$0.020 & 0.165$\pm$.003 \\
    MLP (one layer) & 0.079$\pm$0.004 & 0.137$\pm$0.006 & 0.302$\pm$0.018 & 1.219$\pm$0.022 & 0.162$\pm$0.003 \\
    MLP (three layers) & 0.075$\pm$0.004 & 0.136$\pm$0.006 & 0.305$\pm$0.017 & 1.218$\pm$0.023 & 0.162$\pm$0.003 \\
    MLP (ours) & 0.078$\pm$0.008 & 0.136$\pm$0.011 &0.304$\pm$0.037 & 1.220$\pm$0.043 & 0.162$\pm$0.006 \\
    \hline
  \end{tabular}
  \label{table:Ablation_architecture}
  \end{center}
\end{table*}

\begin{table*}[!t]
  \begin{center}
  \footnotesize
  \captionsetup{font=small}
  \caption{\textbf{Ablation study of $\lambda_{\text{reg}}$.} $\modelname$ shows to be insensitive to this hyper-parameter, and it prefers relatively small $\lambda_{\text{reg}}$.}
  \begin{tabular}{cccccccc}
    \hline
    \textbf{$\lambda_{\text{reg}}$}  & \textbf{ECE} &  \textbf{TL-ECE}  & \textbf{MCE} & \textbf{NLL} & \textbf{Brier}\\
    \hline
    $\lambda_{\text{reg}}=1.0$ & 0.101$\pm$0.006 & 0.161$\pm$0.005 & 0.347$\pm$0.021 & 1.251$\pm$0.027 & 0.164$\pm$0.004 \\
    $\lambda_{\text{reg}}=10^{-1}$ & 0.087$\pm$0.004 & 0.145$\pm$0.005 & 0.318$\pm$0.015 & 1.225$\pm$0.023 & 0.163$\pm$0.003 \\
    $\lambda_{\text{reg}}=10^{-2}$ & 0.078$\pm$0.008 & 0.136$\pm$0.011 & \textbf{0.304$\pm$0.037} & \textbf{1.220$\pm$0.043} & 0.162$\pm$0.006 \\
    $\lambda_{\text{reg}}=10^{-3}$ & \textbf{0.076$\pm$0.004} & \textbf{0.134$\pm$0.006} & 0.311$\pm$0.018 & 1.221$\pm$0.022 & \textbf{0.162$\pm$0.003} \\
    \hline
  \end{tabular}
  \label{table:Ablation_lambda}
  \end{center}
\end{table*}

\paragraph{$\modelname$ Architecture}
To determine if the impressive calibration performance of $\modelname$ is influenced by its specific architectural choice, we conduct an ablation study exploring different model structures. The variants of model architecture include: (1) Linear: a simple model consisting of a single linear layer without a nonlinear activation function; (2) MLP (no hidden layer): A MLP with only input and output layers, excluding any hidden layers; (3) MLP (two hidden layer): A deeper MLP with two hidden layers, offering more complexity. Each of these architectural choices of $\modelname$ is evaluated on the MMLU dataset, the average ECE results are shown in Table~\ref{table:Ablation_architecture}. The ablation results reveal that $\modelname$'s calibration effectiveness is not sensitive to the architecture, as long as the model is sufficiently expressive for calibration of different tasks.

\paragraph{Regularizer Weight} 
To explore the sensitivity of $\modelname$ to hyper-parameter tuning, particularly the regularizer weight $\lambda_{\text{reg}}$ in the training objective, we conduct an ablation study with varying values of $\lambda_{\text{reg}}$. The results shown in Table~\ref{table:Ablation_lambda} demonstrate that while $\modelname$ shows a overall robustness to changes in the regularizer weight, it prefers smaller values of $\lambda_{\text{reg}}$. Intuitively, a larger $\lambda_{\text{reg}}$ upweights the prior term in the training objective and more strongly discourages large temperatures. We also tried a variant with $\lambda_{\text{reg}} = 0$, but this caused numerical instability during training and the resulting models exhibited poor performance. 

\paragraph{Temperature Inference Batch Size at Train Time}
We next explore the sensitivity of $\modelname$ to the size of the batch used to estimate the temperature in training, i.e. $b$ in Algorithm \ref{alg:Alg1}. Based on Lemma \ref{lemm:Conc}, we expect the method to be relatively insensitive to this parameter. Here we use $\lambda_{\text{reg}} = 0.01$ and testing temperature batch size 128.
Results are shown in Table \ref{table:Ablation_train}, confirming this expectation and showing that a batch size of 1 is insufficient.

\begin{table*}[!t]
  \begin{center}
  \footnotesize
  \captionsetup{font=small}
  \caption{\textbf{Ablation study of training temperature inference batch size.} $\modelname$ is seen to be relatively insensitive to this hyper-parameter.}
  \begin{tabular}{cccccccc}
    \hline
    \textbf{Training temperature batch size $b$}  & \textbf{ECE} & \textbf{TL-ECE}   & \textbf{MCE} & \textbf{NLL} & \textbf{Brier}\\
    \hline
    1 & 0.083$\pm$0.009 & 0.140$\pm$0.012 & 0.321$\pm$0.035 & 1.222$\pm$0.045 & 0.162$\pm$0.006 \\
    16 & 0.074$\pm$0.008 & 0.135$\pm$0.012  &  0.291$\pm$0.036 & 1.217$\pm$0.045 & 0.162$\pm$0.006 \\
    32 & 0.077$\pm$0.008 & 0.135$\pm$0.012 & 0.309$\pm$0.035 & 1.220$\pm$0.045 & 0.162$\pm$0.006 \\
    128 & 0.078$\pm$0.008 & 0.136$\pm$0.011 & {0.304$\pm$0.037} & {1.220$\pm$0.043} & 0.162$\pm$0.006 \\
    \hline
  \end{tabular}
  \label{table:Ablation_train}
  \end{center}
\end{table*}

\paragraph{Temperature Inference Batch Size at Test Time}

Similarly, we explore the sensitivity of $\modelname$ to the size of the batch used to estimate the temperature at test time, i.e. $N_\ast$. $N_\ast$ may need to be small in practice if only a few unlabeled examples are available before picking a temperature to run on the task. Based on Lemma \ref{lemm:Conc}, we expect the method to be relatively insensitive to this parameter. Here we use $\lambda_{\text{reg}} = 0.01$ and training temperature batch size 128.
Results are shown in Table \ref{table:Ablation_test}, confirming this expectation.

\begin{table*}[!t]
  \begin{center}
  \footnotesize
  \captionsetup{font=small}
  \caption{\textbf{Ablation study of testing temperature inference batch size.} $\modelname$ is seen to be relatively insensitive to this hyper-parameter.}
  \begin{tabular}{cccccccc}
    \hline
    \textbf{Testing temperature batch size $N_\ast$}  & \textbf{ECE} & \textbf{TL-ECE}   & \textbf{MCE} & \textbf{NLL} & \textbf{Brier}\\
    \hline
    1 & 0.081$\pm$0.009 & 0.138$\pm$0.012 &  0.299$\pm$0.034 & 1.220$\pm$0.045 & 0.162$\pm$0.006 \\
    16 & 0.074$\pm$0.008 &  0.135$\pm$0.011 & 0.296$\pm$0.036 & 1.217$\pm$0.045 & 0.162$\pm$0.006 \\
    128 & 0.078$\pm$0.008 & 0.136$\pm$0.011 &  {0.304$\pm$0.037} & {1.220$\pm$0.043} & 0.162$\pm$0.006 \\
    \hline
  \end{tabular}
  \label{table:Ablation_test}
  \end{center}
\end{table*}

\section{Derivation and Proofs}
\subsection{Proof of Lemma \ref{lemm:Conc}}
\label{app:proof1}
Recall that by construction $\psi_{\param}(\phi(\mathbf{x})) \geq 0$ (desirable since we are estimating a temperature). We can apply a Bernstein inequality to obtain
\begin{equation}
\label{eq:prb}
    \mathrm{Pr}\left(\left|  \sum_{n=1}^{N_\ast} \psi_{\param}(\phi(\feat_n)) - \left(\mathbb{E}\psi_{\param}(\phi(\feat))\right)\right| > t \right) < 2 \mathrm{exp}\left(- \frac{3t^2   }{3N_\ast V_{\theta} + 2C_{\theta} t} \right).
\end{equation}
Let us choose $t$ to achieve inverse squared probability concentration in $N_\ast$, specifically, we set the right hand size of \eqref{eq:prb} to $2N_\ast^{-2}$ and solve for the corresponding $t$:
\begin{align*}
- 2\log N_\ast &:= - \frac{3t_\ast^2   }{3N_\ast V_{\theta} + 2C_{\theta} t_\ast}\\
-6 N_\ast(\log N_\ast) V_{\theta}& - 4(\log N_\ast) C_{\theta} t_\ast + 3t_\ast^2 = 0
\end{align*}
The quadratic theorem yields\footnote{Recall we need $t_{\ast} \geq 0$.}
\begin{align*}
t_{\ast} &= \frac{4(\log N_\ast) C_{\theta} + \sqrt{16(\log N_\ast)^2 C^2_{\theta} + 72V_{\theta} N_\ast \log N_\ast}}{6}\\
&= \frac{2}{3} C_{\theta} \log N_\ast + \sqrt{\frac{4}{9}(\log N_\ast)^2 + 2 V_{\theta} N_\ast \log N_\ast }\\
&\leq \frac{4}{3} C_{\theta} \log N_\ast + (2 V_{\theta})^{\frac{1}{2}} \sqrt{N_\ast \log N_\ast}.
\end{align*}
Substituting this in and dividing both sides by $N_\ast$ yields the first part of the lemma statement.
The last statement in the lemma is then an immediate consequence of the Lipschitz assumption. 

\subsection{Product of Gaussians}
\label{app:proof2}
From standard exponential family properties we have,
\begin{equation}
\prod_{n=1}^{N_k} \gN(x \mid \mu_n, \Sigma_n) \propto \gN(x \mid \hat{\mu}, \hat{\Sigma}),\; \text{where } \hat{\Sigma} = \left(\sum_{n=1}^{N_K} \Sigma_n^{-1} \right)^{-1}\text{and } \hat{\mu} = \hat{\Sigma}\sum_{n=1}^{N_k} \Sigma_n^{-1}\mu_n.
\label{eq:standard}
\end{equation}
\cref{{eq:vapprox}} follows from plugging in $\Sigma_1 = \Sigma_2 = \ldots = \Sigma_{N_k} = \epsilon$ and $\mu_n = \psi_{\param}(\phi(\feat_n^k))$ in \cref{eq:standard}.

\clearpage
\section{Experimental Setup} \label{app:setup}
\subsection{Prompt Templates} \label{app:prompts}
The construction of prompt templates and completion tokens for multiple-choice QA tasks and QA with free form answers is outlined in Table~\ref{table:propmt}. For the multiple-choice QA task, the prompt consists of a question followed by possible answers labeled with letters. The LLM's task is to predict the letter corresponding to the correct answer option. The completion token is the ground-truth label token as provided in the original datasets.

For QA with free form answers, the prompt construction begins with generating a response sequence, denoted as ``RESPONSE,'' based on the given context and question. To limit the length of the LLM's response, we include the prompt ``Answer in as few words as possible'' guiding the LLM to provide concise answers. After generation, this response is concatenated with the original context and question to form a single prompt. This reconstructed prompt is then used to query the LLM on the correctness of its response in a self-checking manner.
Defining the completion token for this task requires a additional step. We employ ROUGE-L precision as the metric to evaluate the correctness of the LLM's generated response. ROUGE-L precision calculates the ratio of the Longest Common Subsequence (LCS) of word sequences to the total number of unique tokens in the target sequence. Considering that the LLM might generate extra tokens not relevant to the actual answer, we consider a response correct as long as the ROUGE-L precision exceeds zero. Given the ROUGE-L based completion token as the groud-truth label token, we measure the accuracy of LLM's self-checking procedure on the six development MRQA datasets in Table~\ref{table:self-check}. Overall, \llamaname~ can accurately predict whether its own response is correct or incorrect, and our proposed $\modelname$ can further calibrate its self-prediction.
\begin{table*}[!t]
  \begin{center}
  \footnotesize
  \captionsetup{font=small}
  \caption{ \textbf{Prompt Template.} ``RESPONSE" denotes the LLM's generated response given the context and questions, and ``TARGET" represents the ground truth answer of the given question. ``ROUGE-L" is a metric to measure the similarity between ``RESPONSE" and ``TARGET.''}
  \begin{tabular}{|c|c|c|}
    \hline
   \textbf{Task}  &  \textbf{Prompt} &  \textbf{Completion Token}\\
    \hline
    \textbf{Multiple-choice QA} &  \makecell{Choose A, B, C, or D.\\ Question: \{``QUESTION"\}\\ A. \{``OPTION 1"\}\\ B. \{``OPTION 2"\}\\ C. \{``OPTION 3"\}\\ D. \{``OPTION 4"\}\\ Answer: \{``MASK" \} } & $\resp_n \in$ \{``A", ``B", ``C", ``D"\} \\
    \hline
    \textbf{QA with Free form Answers}  & \makecell{Choose A, or B.\\ Answer in as few words as possible.\\
    Context: \{``CONTEXT"\} \\ Question: \{``QUESTION"\}\\ Answer: \{``RESPONSE"\}\\ Is the above answer correct? \\A. Yes\\B. No \\ Answer: \{``MASK" \} } & $\begin{cases}\resp_n= \text{``A"}, & \text{If } \text{ROUGE-L(RESPONSE, TARGET)}>0\\ \resp_n=\text{``B"}, & \text{Otherwise} \end{cases}$
                                            \\
    \hline
  \end{tabular}
  \label{table:propmt}
  \end{center}
\end{table*}

\begin{table*}[!t]
  \begin{center}
  \footnotesize
  \captionsetup{font=small}
  \caption{\textbf{Accuracy of \llamaname~'s Self-checking Procedure.} }
  \begin{tabular}{|c|c|c|c|c|c|}
    \hline
    \textbf{BioASQ}	& \textbf{DROP}	& \textbf{DuoRC}	&\textbf{RACE}	&\textbf{RelationExtraction}	&\textbf{TextbookQA}\\
    \hline
    80.6$\%$ &77.7$\%$ &81.6$\%$ &72.0$\%$ &88.7$\%$ &70.5$\%$\\
    \hline
  \end{tabular}
  \label{table:self-check}
  \end{center}
\end{table*}

\clearpage

\begin{table}[!t]
  \centering
  \footnotesize
  \captionsetup{font=small}
  \caption{ \textbf{MMLU Data Examples.}}
  \begin{tabular}{|p{0.20\linewidth} | p{0.55\linewidth}|p{0.15\linewidth}|}
    \hline
   \textbf{Dataset Name}  &  \textbf{Prompt} &  \textbf{Completion Token}\\
    \hline
    \textbf{Anatomy} &  Choose A, B, C, or D. 
    
    Question: The best place to listen to the general heart sound with a stethoscope is the. 
    
    A. fifth left intercostal space in the midclavicular line. 
    
    B. second left intercostal space one inch from the sternum. 
    
    C. third left rib at its junction with the sternum.
    
    D. sternum midway between the sternal angle and xiphisternum. 
    
    Answer: \{``MASK" \} & ``A" \\
    \hline
    
    \textbf{Elementary Mathematics} &  Choose A, B, C, or D. 
    
    Question: Rosa has a goal of running a total of 100 miles this month. Each day that she ran, she ran 5 miles. Which expression could Rosa use to determine how many miles she has left to run after running for d days?
    
    A. $100 - 5d$. 
    
    B. $5d + 100$. 
     
    C. $100 / 5d$.
    
    D. $5d$. 
    
    Answer: \{``MASK" \} & ``A" \\
    
    \hline
    \textbf{Professional Law} & Choose A, B, C, or D. 
    
    Question: Carol Collector was a serious antique car buyer, and was always searching for Thunderbirds in good shape. She saw a newspaper ad offering a 1964 Thunderbird sports coupe for $25,000$, The ad also gave a website address ``to view a picture of the car," which Carol visited, and which showed a T-Bird with a perfect body and interior. Carol paid the $25,000$ and signed a sales contract which specified that the Thunderbird was ``used and sold as is." When Carol went to pick up the car, she learned it was a 1968 model, which was not at all rare and worth much less than the advertised 1964 model. If Carol brings suit against the seller, the likely outcome is for
    
    A. Carol, because the internet picture was of a car that had never been driven, not the actual used car she was to buy. 
    
    B. The seller, because the buyer was aware the Thunderbird was sold ``as is." 
    
    C. The seller, unless they were negligent in not explicitly stating the picture was of a new car, not the actual car to be sold. 
    
    D. Carol, because the ``as is" disclosure was not totally effective. 
    
    Answer: \{``MASK" \} & ``D" \\
    \hline
  \end{tabular}
  \label{table:data-mmlu}
  
\end{table}

\begin{table}[!t]
  \centering
  \footnotesize
  \captionsetup{font=small}
  \caption{ \textbf{BIG-bench Data Examples.}}
  \begin{tabular}{|p{0.20\linewidth} | p{0.55\linewidth}|p{0.15\linewidth}|}
    \hline
   \textbf{Dataset Name}  &  \textbf{Prompt} &  \textbf{Completion Token}\\
    \hline
    \textbf{bbq\textunderscore lite\textunderscore json} &  Choose A, B, or C. 
    
    Question: The parent was considering loaning their car to either their daughter or their elderly parent for the day. Who was not reckless?
    
    A. Cannot be determined. 
    
    B. The daughter. 
    
    C. The elderly parent.
    
    Answer: \{``MASK" \} & ``C" \\
    \hline
    
    \textbf{elementary\textunderscore math\textunderscore qa} &  Choose A, B, C, D, or E. 
    
    Question: What is the result of the following arithmetic operations?:divide 2 by 5, multiply 100 by the result.
    
    A. $32$. 
    
    B. $34$. 
     
    C. $36$.
    
    D. $38$. 

    E. $40$. 
    
    Answer: \{``MASK" \} &  ``E" \\
    
    \hline
    \textbf{fact\textunderscore checker} & Choose A, or B. 
    
    Question: On June 2017, the following claim was made: Mutiny on the Bounty is the second American film based on Treasure Island, the novel. Q: Was this claim true or false?
    
    A. true.
    
    B. false.  
    
    Answer: \{``MASK" \} & ``B" \\
    \hline
    \textbf{goal\textunderscore step\textunderscore wikihow} & Choose A, B, C, or D. 
    
    Question: The most reasonable goal of `Prepare your finances and research the host country before you leave' is
    
    A. Volunteer at a Homeless Shelter.
    
    B. Volunteer with UNHCR.  

    C. Volunteer at a Dog Shelter.

    D. Volunteer at the Local Nursing Home.
    
    Answer: \{``MASK" \} & ``B" \\
    \hline
    \textbf{timedial} &  Choose A, B, or C. 
    
    Question: Which phrase best fits the \{MASK\} span? Context: A: Guess what came in the mail today? B: What? A: My acceptance letter to Yale ! B: Wow ! Congratulation ! When do classes start? A: Freshman orientation is the last week of august, but I want to go \{MASK\} before that to get settled in. B: You're so lucky ! Do you have to do many things before you leave? A: Yes. I'll be very busy ! I have to get a visa, buy a plane ticket, and pack my things. But first, I want to register for classes. B: When can you do that? A: Well, they sent me their prospectus, so I can start looking now. do you want to help me decide which classed to take? B: Sure. What can you choose from? A: Well, I have to take all the fundamental course, plus a few from my major. B: What is your major? A: I hope to major in English literature, but the admissions counselor told me that many people change their major many times in their first year, so we'll see. B: What are the fundamental course? A: In order to graduate, every student must take a certain amount of classes in history, math, English, philosophy, science and art. B: Interesting. That's very different from the Chinese education system. A: Yes, it is. It's also very different from the British education system. B: Really? A: oh, sure. In British, students don't have to take the foundation course. B: why not? A: maybe because they think they know everything already ! ha ha.
    
    A. one day. 
    
    B. one year. 
    
    C. two weeks.
    
    Answer: \{``MASK" \} &  ``C" \\
    \hline
  \end{tabular}
  \label{table:data-bigbench}
\end{table}

\begin{table}[!t]
  \centering
  \footnotesize
  \captionsetup{font=small}
  \caption{ \textbf{MRQA In-domain Train/Validation Data Examples.}}
  \begin{tabular}{|p{0.20\linewidth} | p{0.25\linewidth} |p{0.30\linewidth}|p{0.15\linewidth}|}
    \hline
   \textbf{Dataset Name}  &  \textbf{Prompt} &  \textbf{LLM's Response v.s. True Target } &  \textbf{Completion Token}\\
    \hline
    \textbf{SQuAD} &  Choose A, or B. 

    Answer in as few words as possible.

    Context: Luther's 1541 hymn "Christ unser Herr zum Jordan kam" ("To Jordan came the Christ our Lord") reflects the structure and substance of his questions and answers concerning baptism in the Small Catechism. Luther adopted a preexisting Johann Walter tune associated with a hymnic setting of Psalm 67's prayer for grace; Wolf Heintz's four-part setting of the hymn was used to introduce the Lutheran Reformation in Halle in 1541. Preachers and composers of the 18th century, including J. S. Bach, used this rich hymn as a subject for their own work, although its objective baptismal theology was displaced by more subjective hymns under the influence of late-19th-century Lutheran pietism.

    Question: What is Psalm 67 about?

    Answer: Prayer for grace. 

    Is the above answer correct? 
    
    A. Yes.
    
    B. No. 
    
    Answer: \{``MASK" \} 
    & LLM's Response: Prayer for grace.

    True Target: Prayer for grace; grace.

    ROUGE-L Precision: 0.128.

    &``A" \\
    \hline
    \textbf{NaturalQuestionsShort} &  Choose A, or B. 

    Answer in as few words as possible.

    Context: $<$P$>$ Epidemiology is the study and analysis of the distribution and determinants of health and disease conditions in defined populations . It is the cornerstone of public health , and shapes policy decisions and evidence - based practice by identifying risk factors for disease and targets for preventive healthcare . Epidemiologists help with study design , collection , and statistical analysis of data , amend interpretation and dissemination of results ( including peer review and occasional systematic review ) . Epidemiology has helped develop methodology used in clinical research , public health studies , and , to a lesser extent , basic research in the biological sciences . $<$/P$>$. 

    Question: epidemiologists attempt to explain the link between health and variables such as. 

    Answer: risk factors.

    Is the above answer correct? 
    
    A. Yes.
    
    B. No. 
    
    Answer: \{``MASK" \} 
    & LLM's Response: risk factors.

    True Target: disease conditions in defined populations.

    ROUGE-L Precision: 0.

    &``B" \\ 
    \hline
  \end{tabular}
  \label{table:data-mrqa-train}
\end{table}

\begin{table}[!t]
  \centering
  \footnotesize
  \captionsetup{font=small}
  \caption{ \textbf{MRQA Out-of-domain Test Data Examples.}}
  \begin{tabular}{|p{0.15\linewidth} | p{0.32\linewidth} |p{0.28\linewidth}|p{0.15\linewidth}|}
    \hline
   \textbf{Dataset Name}  &  \textbf{Prompt} &  \textbf{LLM's Response v.s. True Target } &  \textbf{Completion Token}\\
    \hline
    \textbf{BioASQ} &  Choose A, or B. 

    Answer in as few words as possible.

    Context: The large size of spectrin, the flexible protein promoting reversible deformation of red cells, has been an obstacle to elucidating the molecular mechanism of its function. By studying cloned fragments of the repeating unit domain, we have found a correspondence between positions of selected spectrin repeats in a tetramer with their stabilities of folding. Six fragments consisting of two spectrin repeats were selected for study primarily on the basis of the predicted secondary structures of their linker regions. Fragments with a putatively helical linker were more stable to urea- and heat-induced unfolding than those with a putatively nonhelical linker. Two of the less stably folded fragments, human erythroid alpha-spectrin repeats 13 and 14 (HEalpha13,14) and human erythroid beta-spectrin repeats 8 and 9 (HEbeta8,9), are located opposite each other on antiparallel spectrin dimers. At least partial unfolding of these repeats under physiological conditions indicates that they may serve as a hinge. Also less stably folded, the fragment of human erythroid alpha-spectrin repeats 4 and 5 (HEalpha4,5) lies opposite the site of interaction between the partial repeats at the C- and N-terminal ends of beta- and alpha-spectrin, respectively, on the opposing dimer. More stably folded fragments, human erythroid alpha-spectrin repeats 1 and 2 (HEalpha1,2) and human erythroid alpha-spectrin repeats 2 and 3 (HEalpha2,3), lie nearly opposite each other on antiparallel spectrin dimers of a tetramer. These clusterings along the spectrin tetramer of repeats with similar stabilities of folding may have relevance for spectrin function, particularly for its well known flexibility.. 

    Question: Alpha-spectrin and beta-spectrin subunits form parallel or antiparallel heterodimers?

    Answer: Antiparallel.

    Is the above answer correct? 
    
    A. Yes.
    
    B. No. 
    
    Answer: \{``MASK" \} 
    & LLM's Response: Antiparallel.

    True Target: antiparallel.

    ROUGE-L Precision: 1.0

    &``A" \\
    \hline
    
  \end{tabular}
  \label{table:data-mrqa-test}
  \vspace{-2.5em}
\end{table}

\subsection{Data Processing} \label{app:data}
\paragraph{MMLU}
MMLU (Massive Multitask Language Understanding)~\citep{hendrycks2020measuring} is a popular benchmark commonly used to evaluate the depth of knowledge AI models acquire during pre-training, especially under zero-shot and few-shot settings. It contains 57 diverse subjects including STEM, humanities, and social sciences, ranging from basic to professional levels. Given that the datasets within MMLU typically have small training sets, often including only four data points, we utilize the original testing set for training $\modelname$ and utilize the original validation set for validation. As we evaluate the trained $\modelname$ using cross validation, we always evaluate $\modelname$ on the original testing set of the new task. Examples of the processed data, formatted with prompt templates, are detailed in Table~\ref{table:data-mmlu}.
\paragraph{BIG-bench}
BIG-bench (The Beyond the Imitation Game Benchmark)~\citep{srivastava2022beyond}, with its extensive collection of over 200 datasets, contains a wide range of NLP tasks. These include but not limited to multiple-choice question-and-answer (QA), summarization, logical reasoning, and translation tasks. Within this collection, there are approximately 150 datasets corresponding to multiple-choice QA tasks. However, many of them have only a small number of training instances. We first filter out those datasets whose training set size is smaller than 1000, resulting in the selection of 23 datasets for our experiments. For computational reasons, among the selected datasets, we limit the training set size to a maximum of 4000 data instances, and the validation set size is capped at 1000 data instances. The selected datasets are: \textsc{arithmetic, bbq\textunderscore lite\textunderscore json, cifar10\textunderscore classification, contextual\textunderscore parametric\textunderscore knowledge\textunderscore conflicts, color, elementary\textunderscore math\textunderscore qa,  epistemic\textunderscore reasoning, fact\textunderscore checker, formal\textunderscore fallacies\textunderscore syllogisms\textunderscore negation, goal\textunderscore step\textunderscore wikihow, hyperbaton, logical\textunderscore fallacy\textunderscore detection, mnist\textunderscore ascii, movie\textunderscore dialog\textunderscore same\textunderscore or\textunderscore different, play\textunderscore dialog\textunderscore same\textunderscore or\textunderscore different, real\textunderscore or\textunderscore fake\textunderscore text, social\textunderscore iqa, strategyqa, timedial, tracking\textunderscore shuffled\textunderscore objects, vitaminc\textunderscore fact\textunderscore verification, unit\textunderscore conversion, winowhy}.  Examples of the processed data, formatted with prompts templates, are detailed in Table~\ref{table:data-bigbench}. It is noteworthy that the datasets within BIG-bench exhibit large diversity: they differ not only in the number of answer options but also in the type of tasks, ranging from image classification to logical reasoning and mathematical problem-solving. This diversity makes BIG-bench a particularly challenging benchmark, and it demands more robust generalization capability of $\modelname$.

\paragraph{MRQA}
MRQA (Machine Reading for Question Answering)~\citep{fisch2019mrqa} is widely used reading comprehension task with free-form answers. MRQA comprises a total of 18 datasets, which are split into three distinct categories: (1) Training datasets: includes six datasets, namely SQuAD, NewsQA, TriviaQA, SearchQA, HotpotQA, and NaturalQuestions  which are further split into training and in-domain validation data that come from the same data sources as the training datasets, thus they share a similar data distribution with the training datasets. (2) Out-of-Domain development datasets: consists of six datasets, namely BioASQ, DROP, DuoRC, RACE, RelationExtraction, and TextbookQA. These datasets have substantially different data distributions from the training datasets, including variations in passage sources and question styles. (3) Test datasets: consists of another six datasets with only test data used for evaluating the MRQA shared task. Since the test labels are not available, we do not consider this split in our evaluation. To assess $\modelname$, we train and validate the model using the six training datasets and their in-domain validation splits. We evaluate $\modelname$ on the out-of-domain development datasets. Examples of the processed in-domain data and out-of domain are detailed in Table~\ref{table:data-mrqa-train} and Table~\ref{table:data-mrqa-test}, respectively.

\clearpage
\subsection{$\modelname$ Architecture}\label{app:model}
We use a multi-branch MLP  for $\modelname$ inspired from ensemble learning techniques. The structure of $\modelname$ comprises three independent branches with identical architecture, each consisting of a sequence of two linear layers equipped with a ReLU activation function in between. These branches independently process the input, mapping it into lower-dimensional feature representations, each of dimension one. Next, a final linear layer integrates these three one-dimensional features by concatenating them as a three-dimensional feature vector and generates the output. The primary objective behind this architectural choice of $\modelname$ is to improve its ability to generalize across diverse, unseen tasks. We conjectured and early experiments showed that the varied branches can capture distinct feature representations, potentially contributing to better generalization performance. However, as indicated in the ablation study results presented in Table \ref{table:Ablation_architecture}, the performance of $\modelname$ appears to be quite robust to variations in model architecture. More thoroughly comparing $\modelname$ architecture is part of planned future work. 

\subsection{Implementation Details} 
\label{app:implementation}
\begin{table*}[!t]
  \begin{center}
  \footnotesize
  \captionsetup{font=small}
  \caption{\textbf{Hyper-parameter of $\modelname$}.}
  \begin{tabular}{cccccccc}
    \hline
  \textbf{Batch Size} $N_b$ & \textbf{Epochs} ($M$) & \textbf{Checkpoint} ($m$) & \textbf{lr} ($\gamma$) & \textbf{Weight Decay} & $\lambda_{\text{reg}}$ & \textbf{Prior} $\alpha_0$ & \textbf{Prior} $\beta_0$\\
    \hline
   $128$ & $5000$ & $50$ &  $1\times10^{-3}$ & $1\times10^{-4}$ & $1\times10^{-2}$ & $1.25$ & $4.0$\\
    \hline
  \end{tabular}
  \label{table:hyper-parameter}
  \end{center}
\end{table*}
\paragraph{LLMs}
To efficiently train $\modelname$, we employ a strategy of extracting and storing the last layer features from Large Language Models (LLMs) on the local device. This approach significantly reduces the computational cost of LLM inference during the training stage of $\modelname$. For encoder-decoder model FLAN-T5-XL, we configure the maximum source length at 256 tokens and the maximum target length at 128 tokens for MMLU benchmark. For BIG-bench and MRQA which typically involve longer input sequences, the maximum source length is extended to 1024 tokens. For decoder-only model \llamaname~, we set max sequence sequence length to 1024 tokens for all benchmark datasets.
\paragraph{$\modelname$}
In the implementation of $\modelname$, the input dimension of $\modelname$ is set to be 2048 and 4096 for FLAN-T5-XL and \llamaname~, respectively. Correspondingly, the dimensions of the hidden layers in $\modelname$ are set at 256 for FLAN-T5-XL and 512 for \llamaname~. To ensure that the output of $\modelname$ remains positive, a Softplus activation function is adopted. The optimization of $\modelname$ utilizes the AdamW optimizer, and all the hyper-parameter used for training is summarized in Table~\ref{table:hyper-parameter}. The configuration is consistent across experiments conducted on MMLU, BIG-bench, and MRQA. All experiments are
implemented in PyTorch using Tesla V100 GPU with 32 GB memory and Tesla A100 GPU with 40 GB memory.


\end{document}